%% file: paper.tex
\documentclass[dvipsnames]{article} %
\usepackage{apsara}
\usepackage{amssymb}

\usepackage{xspace}
\newcommand{\name}{\xspace}
\usepackage{authblk}

\usepackage{graphicx}
\usepackage{wrapfig}
\usepackage{xcolor}
\definecolor{base_prob}{HTML}{1F77B4}
\definecolor{reproduce_prob}{HTML}{FF7F0E}
\definecolor{r10528_prob}{HTML}{2CA02C}
\definecolor{ori_have}{HTML}{FCDCC3}
\definecolor{from_teacher}{HTML}{8DD3C7}
\definecolor{from_base}{HTML}{B0C4DE}
\definecolor{base_enhance}{HTML}{EBDEEB}

\usepackage[table]{xcolor}
\newcolumntype{Y}{>{\centering\arraybackslash}X}

\usepackage{booktabs}
\usepackage{graphicx}
\usepackage{enumitem}
\usepackage{wrapfig}
\usepackage{algorithm}
\usepackage{algpseudocode}
\definecolor{tablegray}{gray}{0.92}

\usepackage{microtype}
\usepackage{amsmath}
\usepackage{colortbl}
\definecolor{lightgray}{rgb}{0.9,0.9,0.9}
\usepackage{caption}
\usepackage{subcaption}
\usepackage{setspace}
\usepackage{url}
\usepackage{multirow}
\usepackage{colortbl}
\usepackage{tabularx}
\usepackage{blindtext}
\usepackage{pgfplots}
\pgfplotsset{compat=1.18} 
\usepackage{makecell}
\usepackage{tipa}
\usepackage{siunitx}
\usepackage{nicefrac}
\usepackage{tocloft}
\usepackage{listings}
\usepackage[raster,skins]{tcolorbox} %
\usepackage{xltabular}
\usepackage{adjustbox}
\usepackage{xurl}
\usepackage{pifont}
\usepackage{rotating}
\usepackage[normalem]{ulem}
\useunder{\uline}{\ul}{}

\usepackage{hyperref}

\renewcommand{\ghlink}{\href{https://github.com/D2I-ai/dasd-thinking}{Code}}

\newcommand{\blfootnote}[1]{%
  \begingroup
  \renewcommand{\thefootnote}{}%
  \footnotetext[0]{#1}%
  \endgroup
}

\title{Distribution-Aligned Sequence Distillation for \\ Superior Long-CoT Reasoning}

\author{
Shaotian Yan\textsuperscript{*}, Kaiyuan Liu\textsuperscript{*}, Chen Shen\textsuperscript{*,\dag}, Bing Wang\textsuperscript{*}, Sinan Fan\textsuperscript{*}, Jun Zhang, Yue Wu, Zheng Wang, Jieping Ye \\
\textbf{Alibaba Cloud Computing}
}

\begin{document}
\maketitle

\vspace{-1.1cm}
\begin{center}
\begin{tabular}{lll}
\huggingface \ \href{https://huggingface.co/collections/Alibaba-Apsara/dasd-thinking}{Models \& Datasets} \quad &
\modelscope \ \href{https://modelscope.cn/collections/Alibaba-Apsara/DASD-Thinking}{Models \& Datasets} \quad &
\github \ \ghlink \\
\end{tabular}
\end{center}

\blfootnote{\noindent\footnotesize
\textsuperscript{*}Core contributors. \quad
\textsuperscript{\dag}Project lead. \quad
}

\begin{abstract}
\input{sections/0_Abstract} 
\end{abstract}

\vspace{-6pt}
\begin{figure}[h]
  \centering
  \includegraphics[scale=0.275]{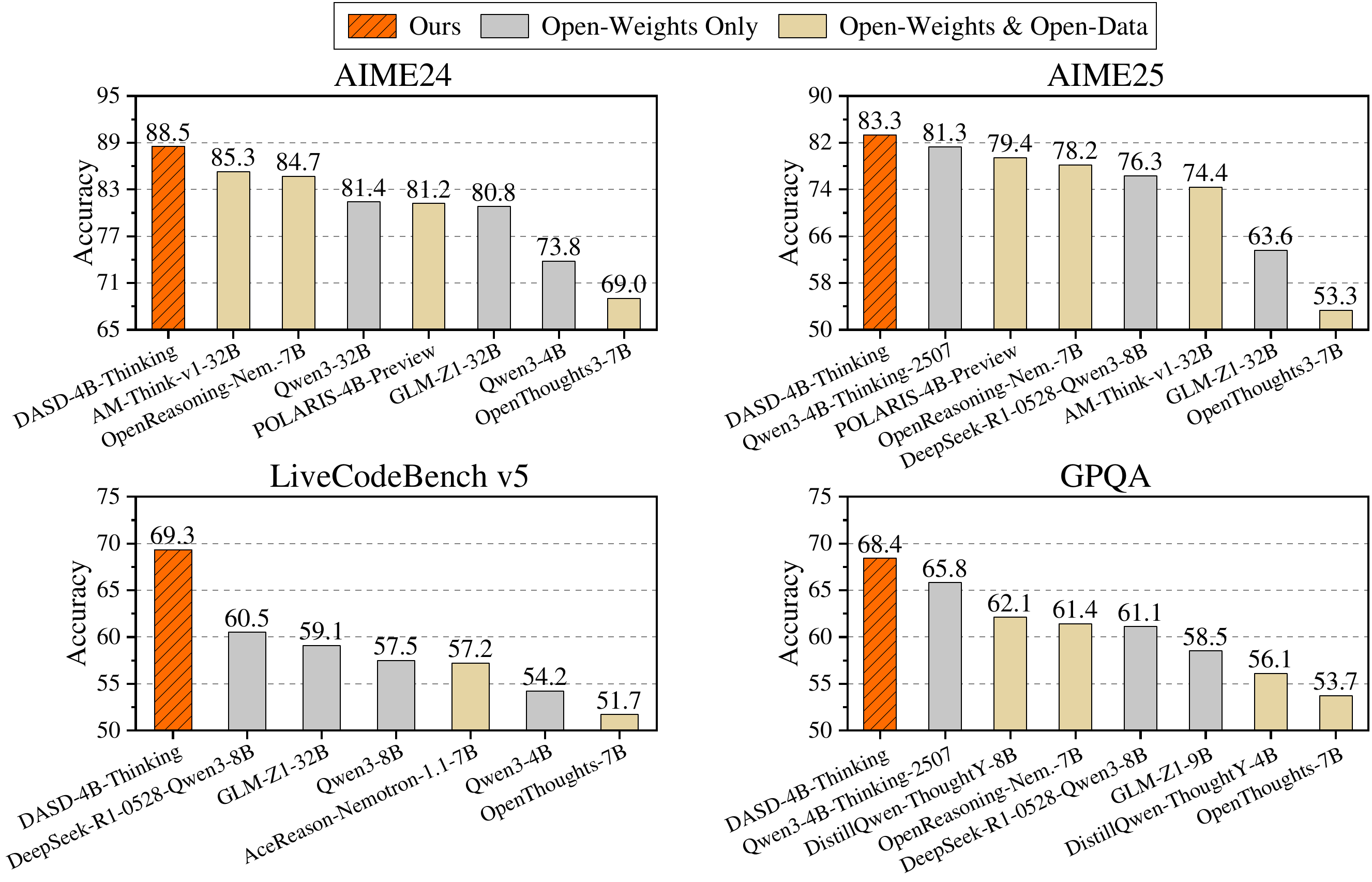}
  \caption{Performance of DASD-4B-Thinking on benchmark datasets. All metrics for the comparison models are taken from their official reports.}
  \label{performamce}
\end{figure}

\newpage

\input{sections/1_Introduction}

\input{sections/2_Discussion}

\input{sections/3_Data}

\input{sections/4_Evaluation}

\input{sections/5_Conclusion}

\bibliography{paper}
\bibliographystyle{apsara}

\end{document}

%% file: sections/0_Abstract.tex
In this report, we introduce \textbf{DASD-4B-Thinking}, a lightweight yet highly capable, fully open-source reasoning model. It achieves state-of-the-art performance among open-source models of comparable scale across challenging benchmarks in mathematics, scientific reasoning, and code generation---even outperforming several larger models (e.g., 32B-scale). We begin by critically reexamining a widely adopted distillation paradigm in the community: supervised fine-tuning (SFT) on teacher-generated responses, also known as sequence-level distillation. Although a series of recent works following this scheme have demonstrated remarkable efficiency and strong empirical performance, they are primarily grounded in the SFT perspective. Consequently, these approaches focus predominantly on designing heuristic rules for SFT data filtering, while largely overlooking the core principle of distillation itself---enabling the student model to learn the teacher’s full output distribution so as to inherit its generalization capability. Specifically, we identify three critical limitations in current practice: i) \textit{Inadequate representation of the teacher’s sequence-level distribution}; ii) \textit{Misalignment between the teacher’s output distribution and the student’s learning capacity}; and iii) \textit{Exposure bias arising from teacher-forced training versus autoregressive inference}. In summary, these shortcomings reflect a systemic absence of explicit teacher–student interaction throughout the distillation process, leaving the essence of distillation underexploited. To address these issues, we propose several methodological innovations that collectively form an enhanced sequence-level distillation training pipeline. Remarkably, DASD-4B-Thinking obtains competitive results using only 448K training samples---an order of magnitude fewer than those employed by most existing open-source efforts. To support community research, we publicly release our models and the training dataset.

%% file: sections/1_Introduction.tex
\section{Introduction}

Recently, DeepSeek \citep{guo2025deepseek} was the first to demonstrate that distillation from powerful teacher models can substantially empower smaller models with reasoning capabilities. Specifically, they curated the reasoning data generated by DeepSeek-R1 and directly fine-tuned several 
widely used open-source compact models. Owing to the simplicity of this supervised fine-tuning (SFT) approach combined with the favorable deployment and inference efficiency of small models, this work has significantly reinvigorated the community’s interest in exploring distillation-based methods for reasoning enhancement.

A series of open-source projects (e.g., OpenR1 \citep{openr1}, OpenThoughts \citep{guha2025openthoughts}, a-m-team \citep{zhao20251}, NVIDIA AceReason \citep{chen2025acereason}, NVIDIA OpenMathReasoning \citep{moshkov2025aimo}, OmniThought \citep{cai2025reasoning}, Light-R1 \citep{wen2025light}, LIMO \citep{ye2025limo}, s1 \citep{muennighoff2025s1}, DeepMath \citep{he2025deepmath}, MiroMind-M1 \citep{li2025miromind}, Syntheic-1 \citep{mattern2025synthetic}, NaturalThoughts \citep{li2025naturalthoughts}, 
and Sky-T1 \citep{novasky2025sky}) have since replicated the DeepSeek-R1 distillation paradigm, dedicating substantial efforts to its faithful reproduction and extension.  
These efforts typically involve collecting and open-sourcing large-scale corpora of challenging reasoning questions, paired with responses generated by powerful teacher models. These datasets have undergone rigorous quality filtering, including stringent correctness verification \citep{lei2025learning,wu2025beyond}, preference-based selection prioritizing higher reasoning difficulty or longer output length \citep{muennighoff2025s1,li2025naturalthoughts}, and diversity-aware curation \citep{jung2025prismatic,li2025exploring}. Subsequently, through SFT on such publicly released reasoning corpora, researchers have obtained distilled models that exhibit strong reasoning capabilities. This paradigm of \textbf{SFT on teacher-generated responses \citep{agarwal2024on}, i.e., sequence-level distillation \citep{kim2016sequence}}, has achieved state-of-the-art or highly competitive performance across diverse domains, including mathematics \citep{openr1, chen2025acereason, wen2025light, he2025deepmath, li2025miromind}, scientific reasoning \citep{guha2025openthoughts, li2025naturalthoughts, mattern2025synthetic}, code generation \citep{ahmad2025opencodereasoning, guha2025openthoughts, zhao20251, cai2025reasoning}, and instruction-following \citep{bercovich2025llama, zhao20251}.

Another paradigm is logit distillation, a classic approach in knowledge distillation \citep{hinton2015distilling} that aligns the logit distributions of student and teacher models to better leverage the rich ``dark knowledge'' encoded in the teacher’s outputs. Notably, recent advancements such as Qwen3 \citep{yang2025qwen3} and Gemma \citep{kamath2025gemma} adopt an on-policy variant: they first generate on-policy sequences using the student model and then align the student’s logit distributions with those of the teacher by minimizing the KL divergence. Recently, Thinking Machines Lab \citep{lu2025on} released an open-source implementation of this paradigm. However, beyond the requirement of accessing token-level logits, these methods face significant challenges when the teacher and student employ different tokenizers, as direct logit alignment becomes infeasible due to misaligned output spaces.

In this work, we aim to improve the aforementioned paradigm---namely, the sequence-level distillation---\textbf{as it is simple and efficient, has already inspired substantial community efforts (including 
the release of a large number of open-source datasets), imposes no arbitrary constraints on the choice of teacher and student model architectures, and does not require access to token-level logits.}
Consistent with \citet{kim2016sequence}, we first argue that SFT on teacher-generated data serves as an effective form of distillation, since such data approximately reflects the teacher model’s sequence-level output distribution and thereby aligns the student model with it.
However, \textbf{existing works in this first paradigm are primarily grounded in the SFT perspective; consequently, they focus predominantly on designing heuristic rules to filter SFT data, while largely overlooking the core principle of distillation itself---enabling the student model to learn the teacher’s full output distribution so as to inherit its generalization capability} \citep{hinton2015distilling}.
In other words, they lack an explicit mechanism to enforce teacher–student interaction throughout the distillation process, leaving the essence of distillation underexploited.
More concretely, such approaches neglect three critical issues:

\begin{itemize}
    \item \textit{From the teacher’s perspective: How to better capture and represent the teacher’s sequence-level distribution?}
    
    \begin{quote}Existing works randomly sample response data under certain quality-based filtering rules. Although such responses can serve as an approximation of the teacher’s sequence-level distribution, this strategy often fails to adequately cover the full support of that distribution. As a result, it may suffer from poor mode coverage or overrepresent low-probability or noisy sequences---making learning particularly challenging for smaller or less capable student models.\end{quote}

    \item \textit{From the student’s learning perspective: How to address misleading gradients when training on teacher-generated data, and more broadly, what target sequence-level distribution better supports effective learning?} 

    \begin{quote}
    Classical knowledge distillation leverages the teacher’s logit distributions to accurately match the student’s predictive distributions over the entire vocabulary at every decoding step, adjusting probabilities up or down as appropriate. 
    In contrast, SFT primarily increases the likelihood of the ground-truth tokens at each prediction position, which can yield misleading gradients (e.g., for tokens the teacher assigns low probabilities but the student assigns high probabilities, SFT pushes those probabilities even higher, driving the student away from the teacher’s distribution). Therefore, identifying a teacher sequence-level distribution that is better aligned with the student model’s learning is of critical importance.\end{quote}
    
    \item \textit{How to mitigate exposure bias caused by training with teacher forcing while evaluating in a free-running, real-world setting?}

    \begin{quote}We emphasize that models distilled on teacher-generated data suffer from pronounced exposure bias: during training, they are exposed to teacher-forced inputs, whereas at inference, they must rely entirely on their own autoregressive predictions. This training–inference mismatch induces a distributional shift, leading to error accumulation and compounding deviations over time. Indeed, we observe that the trained student model’s outputs often diverge from the training distribution in critical aspects---such as response length---and may enter unexpected states that result in incorrect answers.\end{quote}

\end{itemize}

We present a preliminary exploration of approaches to address the above limitations and introduce several key advancements to enhance sequence-level distillation for long chain-of-thought (CoT) reasoning:

\begin{itemize}
    \item \textbf{Temperature-scheduled Learning: Broadening coverage of the teacher’s modes.} 
    A natural approach is to use a higher sampling temperature to better cover the teacher’s output distribution. However, empirical comparison of training convergence across different temperature settings reveals that low-temperature samples exhibit more consistent patterns, making them easier for the student to learn, whereas high-temperature samples---though covering more of the teacher’s modes---introduce greater diversity, hindering learning efficiency. This motivates a temperature-scheduled learning strategy: the student first trains on low-temperature, high-confidence samples to grasp consistent patterns, then gradually incorporates higher-temperature samples to broaden mode coverage. 
    Across the multi-domain settings we evaluated, this two-stage training approach achieves performance gains---particularly in complex reasoning domains such as mathematics and code generation---over single-stage training using either high- or low-temperature sampling. 
    
    \item \textbf{Divergence-aware Sampling: Finding target sequence-level distribution better supports effective learning.}
    Identifying a target distribution---namely, the sequence-level distribution over full responses, as opposed to the token-level logit distribution at each output position---that facilitates effective learning for the student model is nontrivial. Prior work often relies on heuristic rules to select human-expected target distributions, which introduce substantial manual intervention bias and lack theoretical guarantees. To address this, we propose a systematic distribution decomposition framework that analyzes discrepancies between teacher and student predictive probabilities across response candidates. This analysis reveals four canonical distribution patterns. Crucially, we find that one particular pattern---where the teacher assigns high confidence while the student has low probability---consistently correlates with improved test-set performance. 
    Motivated by this finding, we encourage the student model to prioritize learning from such high-divergence instances. Notably, this distribution type naturally mitigates misleading gradients (e.g., from overconfident but incorrect student predictions), and thereby promoting more robust and efficient learning. 

    \item \textbf{Mixed-policy Distillation: Mitigating exposure bias of distilled model.} To mitigate exposure bias, we further introduce a lightweight constructively mixed-policy\footnote{Mixed-policy refers to data generation involving both student and teacher models.} distillation after the initial off-policy SFT phase. Specifically, we randomly select a small subset of training examples, prompt the trained student model to generate full responses, randomly truncate the generated prefixes, and then have the teacher complete the sequence from the truncation point. Only teacher continuations that pass predefined quality filters are retained for the student’s fine-tuning.
    With just a small amount of data and a few additional training steps, this approach yields further performance gains while encouraging more concise model outputs.
\end{itemize}

\begin{wrapfigure}{r}{0.55\textwidth}
  \centering
  \includegraphics[width=0.52\textwidth]{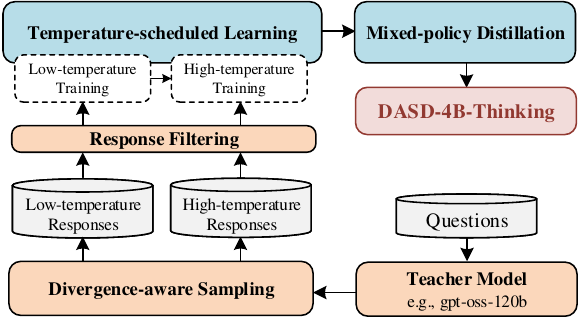}
  \caption{Overall training pipeline of DASD-4B-Thinking.}
  \label{pipeline}
\end{wrapfigure}

Building on these innovations, we present \textbf{DASD-4B-Thinking}, a lightweight yet highly capable reasoning model, post-trained via our
\textbf{D}istribution-\textbf{A}ligned \textbf{S}equence \textbf{D}istillation
pipeline. Specifically, we use Qwen3-4B-Instruct-2507 \citep{yang2025qwen3} as the student model and gpt-oss-120b \citep{agarwal2025gpt} as the teacher model, highlighting the broad compatibility of our approach across diverse model families and architectures. Despite substantial differences between the two models in scale, architecture, vocabulary, tokenizer, and pretraining corpora, our pipeline achieves robust distillation performance. We curate a 
multi-domain dataset spanning mathematics, code generation, scientific reasoning, and complex instruction-following tasks. Figure \ref{pipeline} illustrates the overall training pipeline. We first sample a small amount of cross-domain data at a low temperature and perform one-stage SFT on the student; we then increase the sampling temperature to generate a larger, more diverse training set, resuming training from the checkpoint of the previous stage. Throughout both stages, all synthetic data are generated using our divergence-aware sampling strategy, designed to better align the teacher’s output distribution with the student’s learning capacity. Beyond our core innovations, we also inherit established practices from prior work in this line: our pipeline incorporates rigorous quality control measures, such as filtering truncated outputs and repetitive content, to avoid introducing undesirable patterns into the student model. After these stages, the student exhibits strong reasoning capabilities. To further mitigate exposure bias during autoregressive generation, we introduce a lightweight mixed-policy distillation stage, combining teacher-forced and student-sampled trajectories. This yields the final DASD-4B-Thinking, which balances fidelity to high-quality reasoning paths with robustness against self-generated errors.

DASD-4B-Thinking achieves state-of-the-art performance among models of comparable scale across multiple mainstream reasoning benchmarks in mathematics, code generation, and scientific reasoning---even outperforming some larger models (e.g., 32B-scale). Specifically, it attains 88.5 and 83.3 on the highly challenging mathematical competition benchmarks AIME24 and AIME25, respectively; scores 69.3 on LiveCodeBench v5, a widely adopted benchmark for code generation; and achieves 68.4 on GPQA-Diamond, a doctoral-level scientific reasoning benchmark. Notably, thanks to our methodological innovations, these results are obtained using only 448K training samples---an order of magnitude fewer than those employed by most existing open-source efforts.

We have open-sourced our models (DASD-4B-Thinking and a Mixture-of-Experts (MoE) version DASD-30B-A3B-Thinking-Preview---both derivative models of the Qwen family), together with the training dataset on Hugging Face and ModelScope.

In the following sections, we first elaborate on the motivation, methodology, and roles of the three core components of our approach. Section \ref{section-implementation} details the full training pipeline implementation, and Section \ref{section-evaluaton} presents a comprehensive evaluation of the models.

%% file: sections/2_Discussion.tex
\section{Preliminaries}

The current long CoT distillation paradigm of SFT on teacher-generated data can be traced back to the sequence-level distillation \citep{kim2016sequence}, with the original aim of allowing the student model to mimic the teacher’s distribution at the sequence level and thereby acquiring capabilities comparable to the teacher. Given an input $\boldsymbol{x}$, it trains a student model $p_S$ to minimize the sequence level divergence from the teacher model $p_T$:

\begin{equation}
\label{equ:seq-kl-form}
min D_{KL}(p_T(\boldsymbol{y}\in\mathcal{Y}|\boldsymbol{x})||p_S(\boldsymbol{y}\in\mathcal{Y}|\boldsymbol{x})),
\end{equation}

where $\mathcal{Y}$ is the set of all possible responses that the teacher model can generate for the input prompt $\boldsymbol{x}$. 

This formulation closely parallels classical logit-based distillation \citep{hinton2015distilling}; the key difference is that logit distillation typically matches the teacher’s conditional next-token distribution at each position, whereas sequence-level distillation aligns with the teacher’s distribution over entire output sequences. By modeling the teacher’s distribution, logit distillation conveys the teacher’s implicit ``dark knowledge'', enabling the student to inherit its generalization. Analogously, sequence-level distillation pursues the same objective at the level of complete sequences.

Expanding the KL divergence in Equation \ref{equ:seq-kl-form} yields the following sequence-level distillation objective:

\begin{equation}
    \mathcal{L}_{SEQ} = \sum_{\boldsymbol{y}\in\mathcal{Y}}p_T(\boldsymbol{y} | \boldsymbol{x})[\log p_T(\boldsymbol{y} | \boldsymbol{x}) - \log p_S(\boldsymbol{y} | \boldsymbol{x})],
\end{equation}

the value of $p_T(\boldsymbol{y}|\boldsymbol{x})\log{p_T(\boldsymbol{y}|\boldsymbol{x})}$ depends only on the teacher and is therefore constant with respect to the student’s parameters. It does not affect the gradients and can thus be dropped. Consequently, the sequence-level distillation objective simplifies to:

\begin{equation}
    \mathcal{L}_{SEQ} = -\sum_{\boldsymbol{y}\in\mathcal{Y}}p_T(\boldsymbol{y} | \boldsymbol{x}) \log p_S(\boldsymbol{y} | \boldsymbol{x}).
\end{equation}

However, the exponential size of $\mathcal{Y}$ makes exact computation of $\mathcal{L}_{SEQ}$ intractable. A practical approximation is to use a sampled response $\hat{\boldsymbol{y}}$ and replace $p_T(\cdot\mid \boldsymbol{x})$ with a point mass at $\hat{\boldsymbol{y}}$:
\begin{equation}
\label{equ:seq-approx}
p_T(\boldsymbol{y}\mid \boldsymbol{x}) \approx \mathbb{1}\{\boldsymbol{y} = \hat{\boldsymbol{y}}\}.
\end{equation}

\citet{kim2016sequence} adopt beam search to produce $\hat{\boldsymbol{y}}$, which approximates the mode of $p_T(\cdot | \boldsymbol{x})$ (i.e., the response with the highest probability). In contrast, much of the recent literature on long CoT distillation employs randomly sampled responses as $\hat{\boldsymbol{y}}$. After this approximation, $\mathcal{L}_{SEQ}$ can be further simplified to:
\begin{equation}
\mathcal{L}_{SEQ} \sim -\sum_{\boldsymbol{y}\in\mathcal{Y}}\mathbb{1}\{\boldsymbol{y} = \hat{\boldsymbol{y}}\}\log p_S(\boldsymbol{y} | \boldsymbol{x}) = -\log p_S(\hat{\boldsymbol{y}}|\boldsymbol{x}),
\end{equation}

which exactly recovers the standard SFT loss on teacher-generated outputs. \textbf{This perspective clarifies that the success of SFT on teacher-generated data hinges on effectively transferring the teacher’s sequence-level distribution to the student model}.

Building on the above analysis, \textbf{we argue that the current sequence-level long CoT distillation paradigm should place greater emphasis on teacher–student interaction throughout the distillation process}. However, most existing methods are primarily grounded in the SFT perspective. Consequently, they prioritize filtering high-quality teacher outputs (i.e., SFT data) 
while neglecting such interaction, leading to three main limitations: (i) Inadequate coverage of the teacher’s sequence-level distribution; (ii) Misalignment between the teacher’s output distribution and the student’s learning capacity; and (iii) Exposure bias stemming from teacher forcing during training versus autoregressive inference at test time. In the following sections, we detail our rationale and empirical explorations for the three limitations.

\section{Temperature-scheduled Learning} 
\label{temperature_scheduled_Curriculum}
According to Equation \ref{equ:seq-approx}, SFT on teacher-generated data can implicitly convey the teacher’s distributional information. 
Consequently, the strategy for selecting samples---used to approximate the teacher’s output distribution---plays a pivotal role in how effectively this knowledge is transferred to the student. However, most existing methods for long CoT distillation overlook this consideration, typically relying on random sampling (RS) from the teacher followed by quality-based filtering prior \citep{yan2025towards,lei2025learning}. This approach tends to produce samples that cover only a small subset of the teacher’s modes, thereby under-utilizing the rich latent information embedded in the teacher’s distribution. A natural remedy is to increase the sampling temperature, which flattens the teacher’s distribution and leads to better coverage of its full mode structure \citep{holtzman2019curious,jang2016categorical}. 
\begin{figure}[h]
  \centering
  \includegraphics[width=0.85\textwidth]{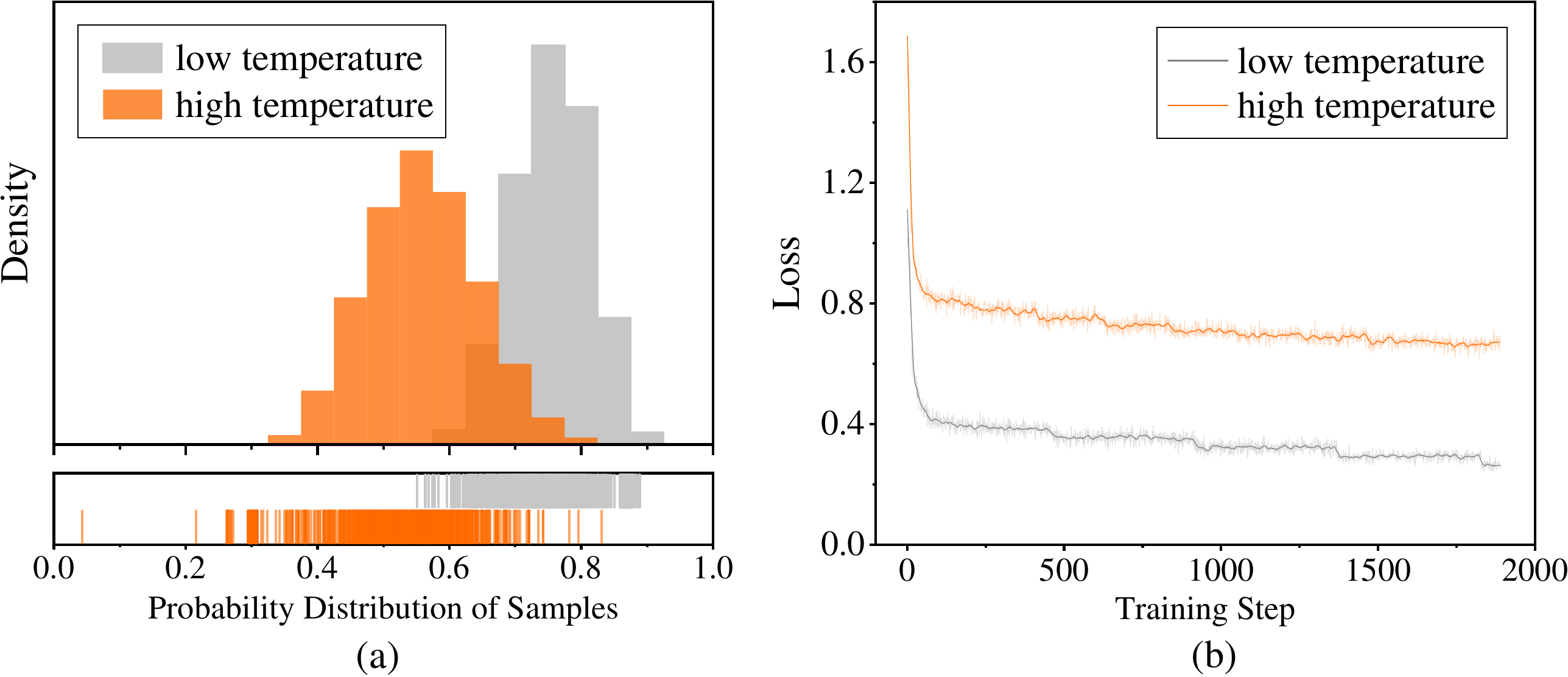}
  \caption{Comparison of probability distribution and training loss with data sampled from \textbf{gpt-oss-120b} under different temperatures. We randomly sampled 50K mathematical reasoning responses at both low (T=0.6) and high (T=1.0) temperatures. To characterize the overall likelihood of a response, we compute \textbf{the geometric mean of its token-level probabilities}.
  (a) Probability distributions of sampled responses: the upper panel displays the density of probability distribution, while the lower panel shows the probability intervals covered by the sampled responses. (b) SFT training loss curves for the student model trained on responses sampled at these temperatures.}
  \label{fig:plot_loss_curve}
\end{figure}

As shown in Figure \ref{fig:plot_loss_curve}(a), we visualize the probability distribution of teacher-generated responses sampled at different temperatures. 
At lower temperatures (T=0.6), the resulting distribution becomes sharper and more peaked, concentrating most probability mass in a narrow range of high-likelihood responses. In contrast, higher-temperature (T=1.0) sampling yields a flatter and broader density, substantially increasing the covered probability range and markedly enhancing data diversity. However, we observe that high-temperature sampling introduces many rare teacher modes or potentially noisy samples. When the student model has limited capacity and exhibits a substantial architectural or behavioral gap from the teacher, it struggles to effectively learn from such heterogeneous data. Figure~\ref{fig:plot_loss_curve}(b) compares the SFT training loss using datasets sampled at different temperatures. Specifically, we randomly sampled 50K math responses from the gpt-oss-120b teacher model at low and high temperatures, and fine-tuned the Qwen3-4B-Instruct-2507 student model on each dataset separately. 
The low-temperature dataset enables rapid convergence to a lower loss with a smooth downward trajectory, whereas the high-temperature dataset makes learning difficult: the loss stays higher.

Despite being more challenging to learn from, training on data sampled at temperature T=1.0 consistently outperforms that sampled at T=0.6, as evidenced in Table~\ref{table_temperature}. Training on T=1.0 samples yields a +1.4 absolute improvement on AIME24 and an even larger +4.2 gain on the more representative and challenging AIME25. This indicates that---even under more difficult optimization dynamics and slower convergence---broader coverage of the teacher’s output modes can lead to substantially greater gains for the student model. It also underscores the crucial impact of the sampling strategy in determining the efficacy of sequence-level distillation. To further evaluate the impact of high-temperature sampling, we scaled the dataset size to 100K samples (doubling the 50K baseline). However, this increase in data volume yields only marginal improvements: as shown in Table~\ref{table_temperature}, the 100K T=1.0 setup achieves no gain on AIME24 relative to 50K, and only a +2.8 improvement on AIME25. This suggests that the student’s capacity to absorb diverse teacher behaviors becomes a bottleneck; adding more high-temperature samples does not translate into proportional performance gains. 
\begin{table}[ht]
\centering
\small
\caption{Performance comparison with different temperature settings.}
\label{table_temperature}
\renewcommand{\arraystretch}{1.3}
\begin{tabularx}{\textwidth}{Xcc}
\toprule
\textbf{Settings of training data} & \textbf{AIME24} & \textbf{AIME25} \\[-2pt]
\midrule
\rowcolor{tablegray} \multicolumn{3}{l}{\textbf{Teacher:} gpt-oss-120b \quad \textbf{Student:} Qwen3-4B-Instruct-2507} \\
50K Math + RS ($T=0.6$) & 81.7 & 71.9 \\
50K Math + RS ($T=1.0$) & 83.1 & 76.1 \\
100K Math + RS ($T=1.0$) & 83.1 & 78.9 \\
50K Math + RS ($T=1.0$) w/ cold start ($T=0.6$) & \textbf{85.2} & \textbf{81.3} \\
\addlinespace[4pt]
\rowcolor{tablegray} \multicolumn{3}{l}{\textbf{Teacher:} Qwen3-Next-80B-A3B-Thinking \quad \textbf{Student:} Qwen3-4B-Instruct-2507} \\
25K Math + RS ($T=0.6$) & 79.0 & 71.3 \\
25K Math + RS ($T=1.0$) & 82.9 & 70.2 \\
25K Math + RS ($T=1.0$) w/ cold start ($T=0.6$) & \textbf{83.1} & \textbf{73.1} \\
\bottomrule
\end{tabularx}
\end{table}

Based on these observations, we propose a temperature-scheduled learning pipeline for sequence-level distillation, an approach inspired by classic logit-based distillation \citep{caron2021emerging, zhou2021ibot}, which we extend to the SFT setting. We begin by sampling from the teacher at a low temperature,
yielding a concentrated set of high-probability, easier-to-learn modes. We then switch to a higher temperature
to collect more diverse samples that capture rarer teacher modes and richer latent information, albeit at the cost of increased learning difficulty. Accordingly, we use the low-temperature data to cold-start the student and then continue training with the high-temperature data. 
As an analogy, this can be intuitively viewed as an easy-to-hard curriculum temperature schedule~\citep{li2023curriculum}, or equivalently, a ``reverse'' version of temperature annealing, a strategy that increases temperature over training, metaphorically inverting the cooling process of conventional annealing.
As shown in Table~\ref{table_temperature}, cold-starting with 50K samples at temperature 0.6 followed by continued training on another 50K
samples at temperature 1.0 yields significant performance gains
over all static-temperature baselines. This demonstrates that our strategy successfully reconciles two objectives: (i) facilitating stable early-stage learning, and (ii) broadening coverage of the teacher’s output distribution, thereby transferring more valuable latent knowledge from the teacher to the student. 

\begin{figure}[h]
  \centering
  \includegraphics[width=0.85\textwidth]{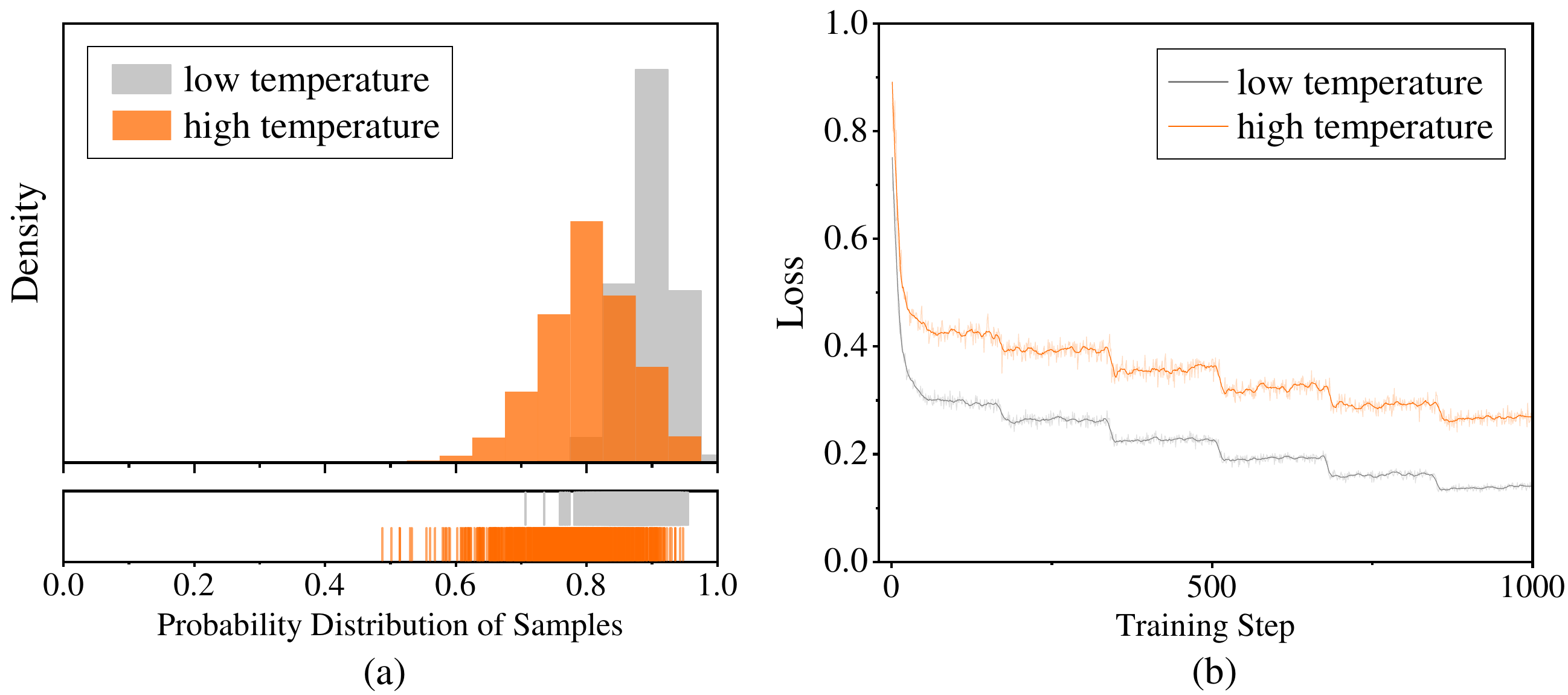}
  \caption{Comparison of probability distribution and training loss with data sampled from \textbf{Qwen3-Next-80B-A3B-Thinking} under different
temperatures.}
  \label{fig:plot_loss_curve_qwen3_next}
\end{figure}

We further validate this approach across a broad range of domains and diverse teacher–student model pairs. In the lower part of Table \ref{table_temperature}, we use Qwen3-Next-80B-A3B-Thinking as the teacher and randomly sample a small-scale of 25K math reasoning traces at temperatures 0.6 and 1.0, respectively.
Continuing training---starting from a cold initialization with 
T=0.6 samples---and incorporating additional T=1.0 samples
yields performance gains of +4.1 on AIME24 and +1.8 on AIME25, which also surpasses training on T=1.0 data alone. As shown in Figure \ref{fig:plot_loss_curve_qwen3_next}, the shift in response probability distributions between T=0.6 and T=1.0 is markedly smaller for Qwen3-Next-80B-A3B-Thinking than for gpt-oss-120b. When training on data from both temperatures, the loss on T=1.0 samples remains higher, but the gap relative to T=0.6 is much narrower than for gpt-oss-120b. Nevertheless, temperature-scheduled learning still improves performance, indicating that samples drawn at different temperatures are complementary. In practice, optimal temperature combinations can be selected based on the model's evaluation performance and the response probability distributions observed across temperature settings. In Table \ref{table_temperature_domain_comparison}, we further evaluate our approach under multi-domain mixed training. Using a small-scale mixture of math, code, and science reasoning data, we show that temperature-scheduled learning remains effective: it delivers substantial gains on AIME25, LiveCodeBench v6, and GPQA Diamond over training with data generated solely at T=0.6 or T=1.0. With the total training data held constant, it also outperforms the T=1.0-only baseline on AIME25 and GPQA Diamond, while achieving comparable performance on LiveCodeBench v6 (potentially attributable to the relatively small proportion of code data in the mixture, where further scaling could still yield noticeable improvements).

\begin{table}[ht]
\centering
\small
\caption{Performance comparison with different settings of training data across domains.}
\label{table_temperature_domain_comparison}
\renewcommand{\arraystretch}{1.3}
\begin{tabularx}{\textwidth}{Xccc}
\toprule
\textbf{Settings of training data} & \textbf{AIME25} & \textbf{LCB v6} & \textbf{GPQA-D} \\[-2pt]
\midrule
\rowcolor{tablegray} \multicolumn{4}{l}{\textbf{Teacher:} gpt-oss-120b \quad \textbf{Student:} Qwen3-4B-Instruct-2507} \\
25K Math + 10K Code + 10K Science + RS ($T=0.6$) & 74.6 & 44.1 & 65.5 \\
25K Math + 10K Code + 10K Science + RS ($T=1.0$) & 75.2 & 47.3 & 65.4 \\
50K Math + 20K Code + 20K Science + RS ($T=1.0$) & 75.8 & \textbf{51.3} & 65.4 \\
25K Math + 10K Code + 10K Science + RS ($T=1.0$) w/ cold start ($T=0.6$) & \textbf{77.5} & 51.0 & \textbf{66.4} \\
\bottomrule
\end{tabularx}
\end{table}

\section{Divergence-aware Sampling}
\label{divergence_aware_sampling}

Despite employing temperature-scheduled learning to broaden coverage of the teacher’s modes, the student still struggles to align with the teacher’s sequence-level distribution. Classical logit distillation leverages teacher logit distribution to precisely calibrate the student’s token-level probabilities, increasing or decreasing them as needed \citep{gu2024minillm,agarwal2024on}. By contrast, SFT on teacher-generated data typically amplifies the probabilities of all target tokens relative to the student’s current predictions. This can induce misleading gradients: for tokens assigned low probabilities by the teacher but high probabilities by the student, SFT erroneously pushes the student’s probabilities even higher, thereby driving them away from the teacher’s distribution. This discrepancy motivates a core question: How can we identify a teacher-derived sequence-level distribution that is better aligned with the student model’s learning capacity?

\begin{figure}[tb]
  \centering
  \includegraphics[width=0.97\textwidth]{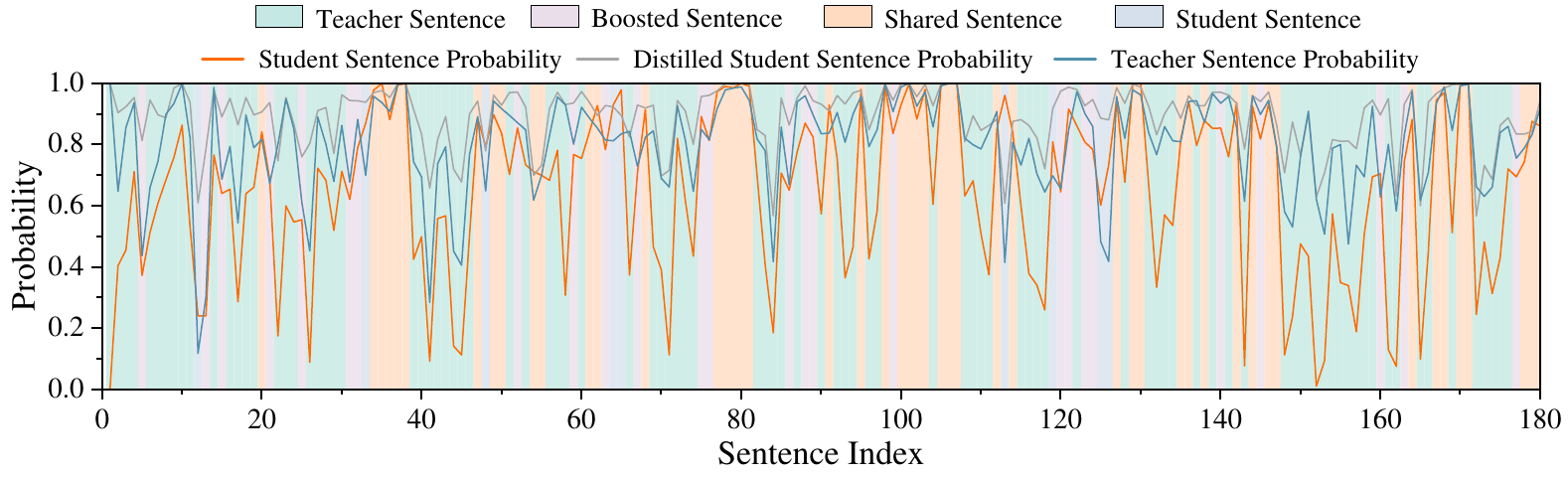}
  \caption{Joint comparison of the three models’ predicted probabilities. An example of output probabilities: the x-axis indexes sentences, and the y-axis shows predicted probabilities. Foreground lines plot the probabilities of the three models, while background colors indicate the inferred source of each sentence. By comparing probability differences, every sentence is categorized into one of four source types.}
  \label{das_fig_action_type_example}
\end{figure}

To identify an effective sequence-level target distribution from the student’s perspective, we introduce a distribution decomposition and analysis framework~\citep{sentence_come_from}: each sequence-level response is decomposed into consecutive sentences and the corresponding sentence-level generation probabilities are computed for both the teacher and the student; by quantifying the probability discrepancy on each shared sentence, we categorize distinct behavioral patterns; finally, we systematically analyze these patterns (i.e., components) and establish their empirical relationship to effective student learning.

Concretely, following the experimental setup in Section~\ref{temperature_scheduled_Curriculum}, we first sample responses from the distilled model (i.e., the trained student model) on test-set prompts and segment each response into sentences. \textbf{This sentence-level analysis ensures the broad applicability of our method across heterogeneous model families}---unlike approaches such as on-policy distillation, which typically requires all models to share the same tokenizer and vocabulary (a constraint imposed by its reliance on token-level supervision).
Then, we feed these samples to the teacher model, the pre-distillation student model (hereafter, the ``student model''), and the post-distillation student model (hereafter, the ``distilled model''). 
For each sentence of the response data, we compute its probability under each of the three models as the geometric mean of per-token probabilities in this sentence.
As illustrated in Figure \ref{das_fig_action_type_example}, we observe that the sequence-level distribution admits a natural decomposition into four well-defined distribution types (each corresponding to a distinct sentence category). Let $p_T$, $p_S$ and $p_D$ denote the predicted probabilities of the teacher, student, and distilled models for the same sentence, respectively. Based on the relative magnitude discrepancies of these probabilities, we define the following distribution (or sentence) types:

\begin{itemize}

\item Student-originated sentences (hereafter referred to as \colorbox{from_base!70}{\textcolor{black}{Student Sentence}}) and teacher-originated sentences (hereafter referred to as \colorbox{from_teacher!50}{\textcolor{black}{Teacher Sentence}}): When there is a large discrepancy between $p_S$ and $p_T$, the distilled model still outputs the sentence, suggesting the sentence is more consistent with the model assigning the higher likelihood. For example, if $p_{T} \gg p_{S}$ and distilled model nevertheless produces the sentence, it is more likely teacher-originated. \textbf{Moreover, when $p_{T} \gg p_{S}$, the student can relatively freely increase its probability under SFT without concern about misleading gradients. Intuitively, this type of pattern is more likely to apply under our current distillation setup.} Note that a Teacher Sentence does not imply that the action is entirely absent from the student model, but rather that it is primarily originated from the teacher. The same applies to a Student Sentence.

\item Pre-existing sentences in both pre-distillation student model and teacher model, not enhanced by distillation (hereafter referred to as \colorbox{ori_have!70}{\textcolor{black}{Shared Sentence}}): The output probabilities for these sentences are similar across all three models. This indicates that these sentences are already well-supported by both the pre-distillation student and the teacher, and that distillation does not materially change their probabilities or increase inter-model distribution discrepancies.

\item Pre-existing sentences boosted through distillation (hereafter referred to as \colorbox{base_enhance!70}{\textcolor{black}{Boosted Sentence}}): Similar to the second type, $p_T$ and $p_S$ remain close, but $p_D$ differs significantly (and $p_D$ is typically higher in practice, since trajectories are sampled from the distilled model). These sentences also exist in both the teacher and the student before distillation, but their probabilities are amplified by training on distilled data.

\end{itemize}

\begin{figure}[t]
  \centering
  \includegraphics[width=1.0\textwidth]{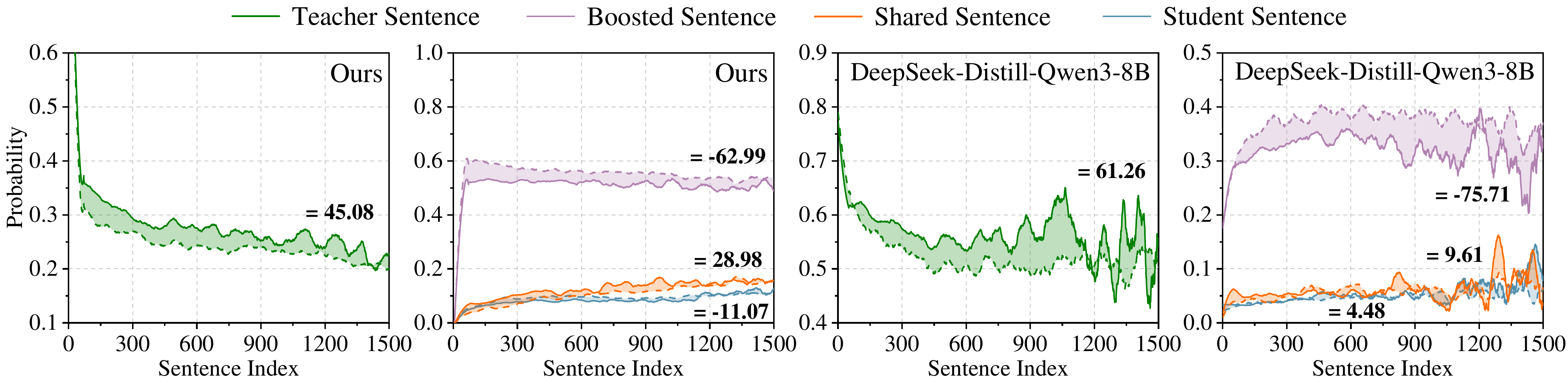}
  \caption{Position-wise distribution over the four sentence types for our internally trained model (left two panels) and the open-source DeepSeek-Distill-Qwen3-8B (right two panels). \textbf{The x-axis denotes the sentence position, and the y-axis denotes the predicted probabilities of the four sentence types. Solid lines (-----) indicate probabilities when the answer is correct, while dashed lines(- - -) indicate probabilities when the answer is incorrect.} $\Delta$ denotes the area difference between the solid and dashed curves, which reflects the influence of each sentence type on answer correctness.}
  \label{fig_insight2_output_source}
\end{figure}

Having decoupled the output distributions, we next investigate which distribution types are most conducive to the student model’s learning (i.e., those that best support effective knowledge acquisition). To this end, we assess effective learning by analyzing the correlation between the four distribution types and test-set answer correctness.
Specifically, for each sentence position, we compute the probability that the distilled model assigns to each distribution type. Since solutions often contain multiple sentences (correct answers typically contain fewer sentences than incorrect ones), analyzing at the sentence-position-level, rather than at the full solution level, allows us to focus more directly on the distribution types themselves and mitigate confounding effects arising from sentence position. For example, to estimate the probability of the Teacher Sentence at the third sentence position, we calculate the fraction of third sentences that are categorized as Teacher Sentence, across all correct and incorrect model outputs. Notably, the number of sentences per answer varies, limiting data availability at later positions.
To ensure statistical reliability, we therefore focus primarily on earlier sentence positions, where sufficient samples exist. 
We also replicate this analysis on the open-source model DeepSeek-Distill-Qwen3-8B \citep{guo2025deepseek} to ensure generalizability. 
\textbf{As shown in Figure \ref{fig_insight2_output_source}, across models, Teacher Sentences tend to receive higher probabilities in correct answers}, evidenced by the light-green solid line (\textcolor{from_teacher}{-----}) persistently lying above the light-green dashed line (\textcolor{from_teacher}{- - -}). This is as expected: since the teacher model performs better on the test set, aligning the student’s outputs with teacher-preferred responses enhances learning efficacy and, consequently, the likelihood of generating correct answers.
In contrast, we find that Shared Sentence and Student Sentence occur with low probability and exert a relatively minor influence. For Boosted Sentence, we observe a potential negative correlation between Boosted Sentences and test-set accuracy. We conjecture that this possibly stem from suboptimal misleading gradients. More importantly, the distillation pipeline only admits the teacher and student models prior to training, rendering it impossible to directly identify Boosted Sentences. We therefore focus primarily on Teacher Sentences in the remainder of this work.

Building on the above analysis, a natural idea is to emphasize, during training, patterns that are more indicative of answer correctness. Although the full distribution-decomposition framework requires output probabilities from three models (the teacher, the student, and the distilled model) to identify the most effective distribution post hoc, we show that Teacher Sentences/Student Sentences can be identified prior to training: Teacher Sentences/Student Sentences are those sentences for which the teacher assigns significantly higher/lower output probabilities than the student. Therefore, we propose divergence-aware sampling (DAS), which prioritizes training examples rich in Teacher Sentences and thereby implicitly targets a teacher-derived sequence-level distribution better aligned with the student’s learning capacity~\citep{sentence_come_from}. 
This sampling distribution naturally mitigates misleading gradients and facilitates more effective knowledge transfer from teacher to student.
\textbf{Notably, our method only requires, for each token in the teacher-generated response, its predicted probability by both the teacher and the student}. The teacher-side probabilities are naturally obtained during sampling---and are often exposed even by many closed-source APIs---while the student-side probabilities are readily computed from the local model. In contrast, classical logit-based distillation necessitates the teacher’s full-vocabulary logits (i.e., probabilities over the entire vocabulary) at every position. \textbf{Even recent on-policy distillation methods---when simplified to operate on token-level probabilities---still require, for every token in the student’s generated outputs, the corresponding probabilities under both models. Critically, the teacher-side probabilities for the student’s outputs are typically unavailable for proprietary models}.

\begin{table}[ht]
\centering
\small
\caption{Performance comparison with different settings of training data (RS vs. DAS).}
\label{table_insight2_1}
\renewcommand{\arraystretch}{1.3}
\begin{tabularx}{\textwidth}{Xcc}
\toprule
\textbf{Settings of training data} & \textbf{AIME24} & \textbf{AIME25} \\[-2pt]
\midrule

\rowcolor{tablegray} \multicolumn{3}{l}{\textbf{Teacher:} gpt-oss-120b \quad \textbf{Student:} Qwen3-4B-Instruct-2507} \\
50K Math + RS ($T=0.6$)             & 81.7 & 71.9 \\
50K Math + DAS ($T=0.6$)            & \textbf{83.3} & \textbf{74.2} \\
\addlinespace[4pt]
50K Math + RS ($T=1.0$)             & 83.1 & 76.1 \\
100K Math + RS ($T=1.0$)            & 83.1 & 78.9 \\
50K Math + DAS ($T=1.0$)            & \textbf{85.0} & \textbf{79.2} \\

\midrule

\rowcolor{tablegray} \multicolumn{3}{l}{\textbf{Teacher:} Qwen3-Next-80B-A3B-Thinking \quad \textbf{Student:} Qwen3-4B-Instruct-2507} \\
25K Math + RS                       & 79.0 & 71.3 \\
25K Math + DAS                      & \textbf{82.5} & \textbf{71.9} \\

\bottomrule
\end{tabularx}
\end{table}

Building on the experimental setup in Section~\ref{temperature_scheduled_Curriculum}, we conduct a controlled comparison between DAS and random sampling (DS) under an identical sampling budget. As shown in Table \ref{table_insight2_1}, DAS consistently achieves higher test performance, and in several cases, even surpasses the results obtained by RS after scaling up its data volume. This demonstrates that DAS effectively identifies teacher-generated sequences whose distribution is better aligned with the student’s learning capacity. Further, as shown in the lower part of Table \ref{table_insight2_1} and in Table \ref{table_insight2_1_different_domain_unified}, DAS maintains a clear advantage over random sampling across different teacher models and domains, validating the generalizability of the DAS method.

\begin{table}[t]
\centering
\small
\caption{Performance comparison with different settings of training data (RS vs. DAS across domains).}
\label{table_insight2_1_different_domain_unified}
\renewcommand{\arraystretch}{1.3}
\begin{tabularx}{\textwidth}{Xccc}
\toprule
\textbf{Settings of training data} & \textbf{AIME25} & \textbf{LCB v6} & \textbf{GPQA-D} \\[-2pt]
\midrule

\rowcolor{tablegray} \multicolumn{4}{l}{\textbf{Teacher:} gpt-oss-120b \quad \textbf{Student:} Qwen3-4B-Instruct-2507} \\

25K Math + 10K Code + 10K Science + RS  & 74.6 & 44.1 & 65.5 \\
25K Math + 10K Code + 10K Science + DAS & \textbf{75.6} & \textbf{47.3} & \textbf{65.7} \\

\bottomrule
\end{tabularx}
\end{table}

Finally, DAS does not require re-sampling data for every new student model. For instance, as demonstrated in Section~\ref{section-evaluaton}, data curated to match the learning capacity of the Qwen3-4B-Instruct-2507 student model generalizes effectively to the Qwen3-30B-A3B-Instruct-2507 student model.

\section{Mixed-policy Distillation}
\label{mixed_policy}

\begin{wrapfigure}{r}{0.44\textwidth}
  \centering
  \includegraphics[width=0.44\textwidth]{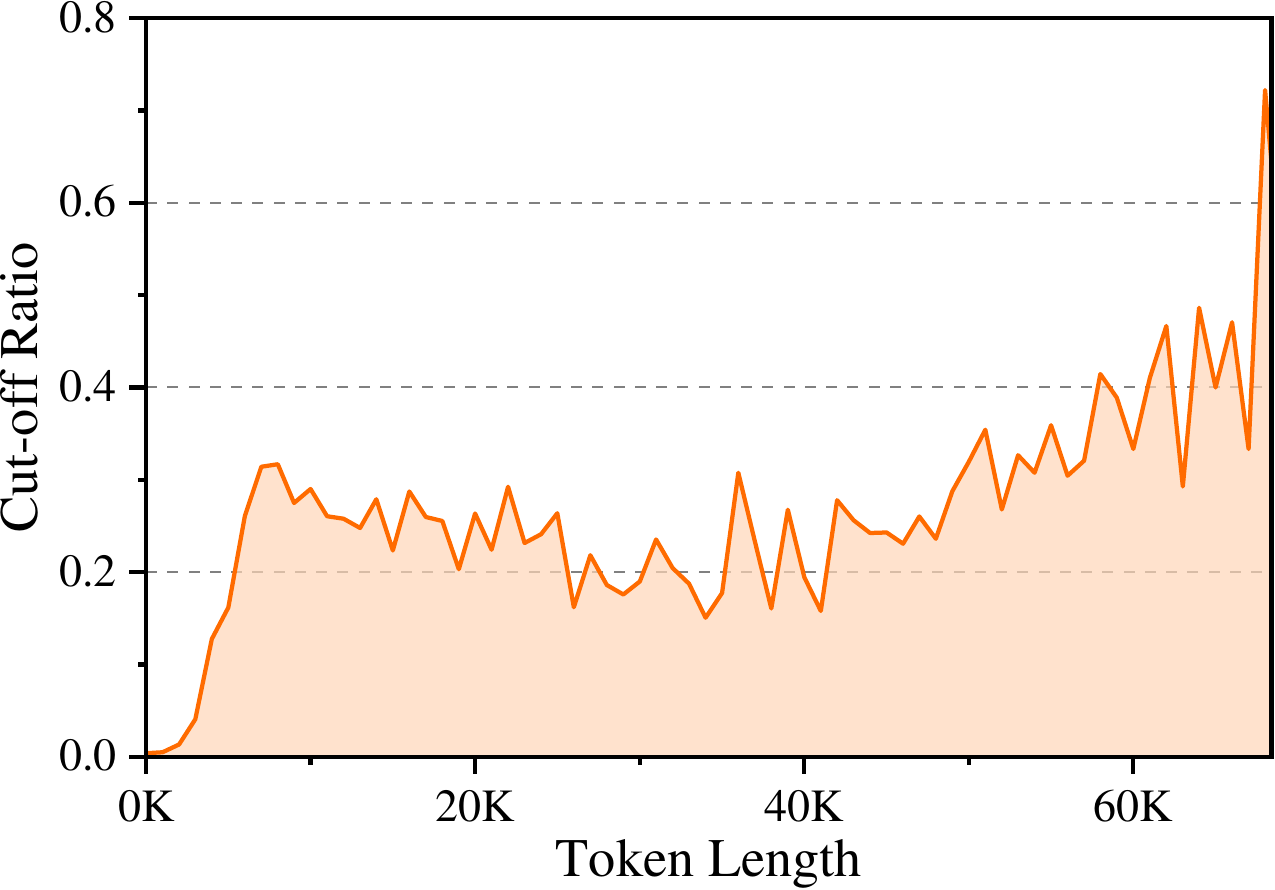}
  \caption{The ratio between cut-off responses under different token lengths.}
  \label{fig:length_cut_ratio}
\end{wrapfigure}

In the previous stages, we employed off-policy methods to approximate the teacher’s sequence-level distribution through high-quality data generation. Nevertheless, we find that the resulting student model still suffers from exposure bias \citep{ranzato2016sequence}: during training, the student is conditioned on the teacher’s prefix using teacher forcing, whereas at inference time, it must rely on its own autoregressive predictions, leading to a distribution mismatch.

To empirically investigate this phenomenon, we use the student model trained in the previous round (50K DAS sampling, T=0.6; see Table~\ref{table_insight2_1}) to re-generate the training data within its own context, in order to examine whether the student model is overly reliant on the teacher’s context.
During inference, we set the maximum generation token length to 1.5 times the length of the teacher-provided solution, in order to compare the differences between the teacher’s reference response and the student’s self-generated counterpart. Figure~\ref{fig:length_cut_ratio} plots the cut-off rate of the student’s generated responses across different training-response lengths, where a higher cut-off rate indicates greater divergence between student and teacher behavior. 
The results reveal that, even on the training data, the student still exhibits substantial deviations from the teacher, and this discrepancy becomes increasingly pronounced as the length of the training response grows. This observation confirms that training with teacher forcing under longer teacher prefixes exacerbates exposure bias.

To overcome these limitations, \citet{chen2025retaining} present that on-policy data collection is an effective alternative method. Accordingly, we propose a mixed-policy distillation approach that synergistically combines off-policy and on-policy signals. Specifically, we first use the student model from the previous training round to re-generate responses for the training queries, and then identify instances that differ substantially from the teacher’s outputs, e.g., solutions that have been cut off in Figure~\ref{fig:length_cut_ratio}. For these data points, we randomly cut off the solutions generated by the student and prompt the teacher to continue the generation, thereby enabling the teacher to provide targeted guidance on the student’s errors.

We present an ablation study of the proposed mixed-policy distillation in Table~\ref{table:rft.ablation}. As described above, we collect 7.7K mixed-policy samples, and train the model with only one epoch across these samples. 
Our baseline is the model trained 
on the 50K DAS-generated dataset at temperature T=0.6 (Table~\ref{table_insight2_1}).
In addition, we investigate a masking variant, where student-generated portions are masked out, and only teacher-completed segments are retained for training.

\begin{wraptable}{r}{0.45\textwidth}
\centering
\small
\vspace*{-1\baselineskip}
\caption{Ablation of the mixed-policy distillation method. \#Num: number of mixed-policy data.}
\label{table:rft.ablation}
\renewcommand{\arraystretch}{1.2}
\setlength{\tabcolsep}{4pt}
\begin{tabularx}{\linewidth}{ccYY}
\toprule
\rowcolor{tablegray} \textbf{\#Num} & \textbf{Mask} & \textbf{AIME24} & \textbf{AIME25} \\
\midrule
\multicolumn{4}{l}{\textit{Baseline}} \\
50K DAS ($T=0.6$) & --- & 83.3 & 74.2 \\
\midrule
\multicolumn{4}{l}{\textit{Mixed-Policy Variants}} \\
7.7K & \ding{51} & 80.8 & 72.3 \\
7.7K & \ding{55} & \textbf{83.3} & \textbf{74.8} \\
\bottomrule
\end{tabularx}
\vspace*{-1\baselineskip}
\end{wraptable}

To maintain a balanced proportion between mixed-policy and off-policy data during training, we introduce 20K additional off-policy samples for joint training with the mixed-policy data.
Our main experimental observations are as follows:
(1) The results indicate that our mixed-policy dataset, despite containing only 7.7K samples, is capable of enhancing the model’s performance.
(2) The masking variant tends to yield worse performance. This is because masking removes the on-policy segments generated by the student, leaving only the off-policy segments from the teacher for training. The observed performance drop in this setting further demonstrates the importance of incorporating on-policy data during distillation.
As shown in Table~\ref{tab: Ablation-results}, we also validate the effectiveness of the mixed-policy distillation approach in our final training pipeline. 
Incorporating even a small amount of mixed-policy data yields measurable gains across strong models and diverse domains.
These results motivate continued exploration of this promising direction in the future.

%% file: sections/3_Data.tex
\section{Overall Training Recipe}
\label{section-implementation}

In this section, we detail the concrete implementation of DASD-4B-Thinking. Our pipeline comprises (i) question collection, (ii) candidate response sampling, (iii) filtering to remove low-quality responses, and (iv) multi-stage training on the curated dataset. Beyond our core innovations, we also integrate well-established practices from prior work in this line to ensure the overall quality of responses. We present each component in turn and describe the associated design choices in detail below.

\subsection{Question Collection}

Our goal is to collect challenging questions spanning a diverse set of domains in order to obtain responses that demonstrate meaningful reasoning. To this end, we select four representative domains: mathematical reasoning, code generation, scientific reasoning, and instruction following. To gather training questions, we utilize a variety of publicly available open-source datasets \citep{chen2025acereason,sci_dataset_OpenScienceReasoning2,ji2025thinking}, which summarize numerous high-quality questions.

\begin{itemize}

\item \textbf{Mathematical Reasoning.} Our math questions are primarily sourced from the supervised fine-tuning dataset of NVIDIA AceReason \citep{chen2025acereason}. 
We experimented with different question scales during training and, considering the diminishing returns with larger question scales, ultimately sampled about 105K questions. 
These questions include those from original sources such as NuminaMath-CoT \citep{math_dataset_numina_math_datasets} and the Art of Problem Solving (AoPS) community forums.

\item \textbf{Code Generation.} Our code questions primarily come from the OpenCodeReasoning dataset \citep{ahmad2025opencodereasoning}, with data sources including TACO \citep{code_dataset_taco}, CodeContests \citep{code_dataset_code_contest}, APPs \citep{code_dataset_apps}, and Codeforces.

\item \textbf{Scientific Reasoning.} Scientific questions are primarily collected from NVIDIA's OpenScience Reasoning dataset \citep{sci_dataset_OpenScienceReasoning2}, which consists entirely of multiple-choice question-answer pairs. We prioritize questions with longer example answers, as they tend to better elicit the model's reasoning abilities.

\item \textbf{Instruction Following.} Instruction-following questions are sourced from AM-DeepSeek-R1-Distilled-1.4M \citep{ji2025thinking}, which contain a large number of subjective instructions from various datasets.

\end{itemize}

\subsection{Response Sampling}

We select a representative student-teacher pair, Qwen3-4B-Instruct-2507 and gpt-oss-120b, as our primary model pair, highlighting the broad compatibility of our approach across diverse model families and architectures.

During sampling, we leverage the teacher’s high-level reasoning capabilities to generate multiple candidate samples for each question. To enhance coverage of the teacher model's behavior, we follow the methodology outlined in Section \ref{temperature_scheduled_Curriculum}, sampling multiple responses per question at both low and high temperatures. Furthermore, we apply divergence-aware sampling as described in Section \ref{divergence_aware_sampling} to prioritize examples that better support student learning while preserving the correct gradient directions. 

\begin{figure}[h]
  \centering
  \includegraphics[width=1.0\textwidth]{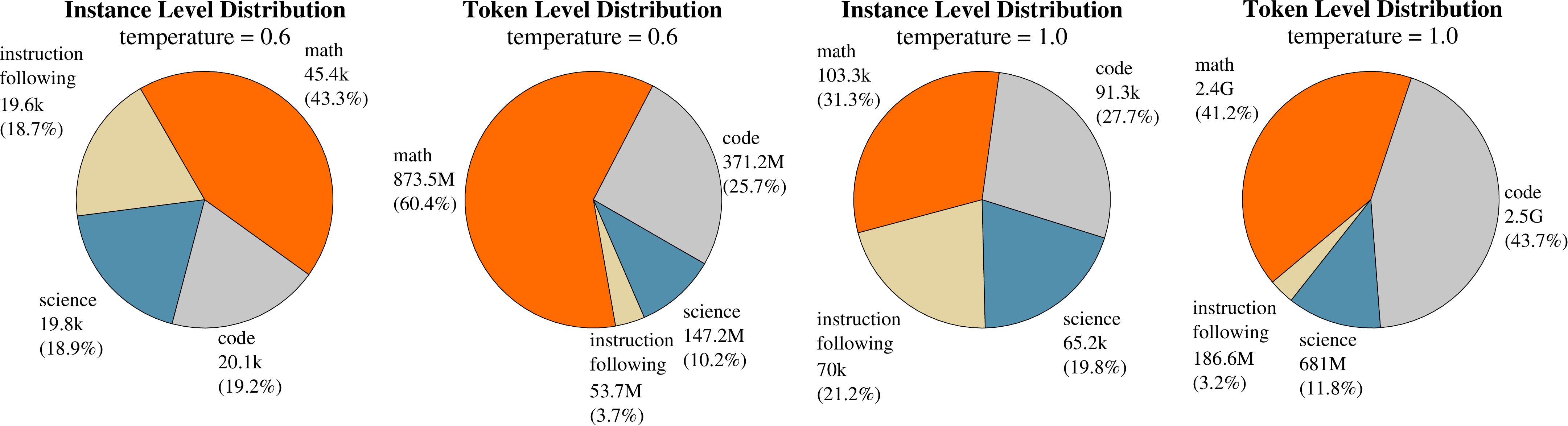}
  \caption{Data distribution.}
  \label{fig_data_distribution}
\end{figure}

\subsection{Response Filtering}

Beyond our core innovations, we employ stringent response filtering strategies to remove low-quality samples:

\begin{itemize}
    \item \textbf{Length-based Filtering.} We compute response lengths using the student model’s tokenizer, whose segmentation may differ substantially from the teacher’s. Responses that exceed the training context length are discarded.
    \item \textbf{Structure-based Filtering.} gpt-oss-120b occasionally invokes built-in tools to answer questions. Since our current focus is on distilling long CoT reasoning capabilities---and function calling is deferred to future work---we explicitly detect and filter out any responses containing function calls. Additionally, we require every response to include both a thinking process and a final answer. gpt-oss-120b uses a harmony template to separate the reasoning trace from the final answer; for student-model compatibility, we post-process the original outputs by replacing the native delimiters with ``$<think>$'' and ``$</think>$''.     
    \item \textbf{Repetitive Content Filtering.} We observe that gpt-oss-120b tends to generate repetitive content, particularly at lower temperatures, including repeated paragraphs, sentences, or phrases within a single response. Such repetitions can induce the trained student to produce endlessly repetitive and excessively verbose outputs during inference. To mitigate this, we rigorously filter out data containing repetitive content using regular expressions and n-gram matching, preventing the student from internalizing undesirable patterns.
\end{itemize}

Finally, we acquire a total of 105K low-temperature (T=0.6) and 330K high-temperature (T=1.0) responses. The resulting data distribution across various domains is shown in Figure~\ref{fig_data_distribution}.

\subsection{Multi-stage Training}
As illustrated in Figure \ref{pipeline}, our training pipeline comprises two main stages: temperature-scheduled learning and mixed-policy distillation. During temperature-scheduled learning, the sampling data used to train DASD-4B-Thinking undergoes a two-stage filtering and training process to better capture the teacher model’s distribution while effectively supporting the student model’s learning. In the subsequent mixed-policy distillation, we construct mixed-policy data via on-policy rejection sampling and off-policy teacher revision. This hybrid strategy mitigates exposure bias by providing targeted, error-aware supervision.

\subsubsection{Temperature-scheduled Learning}

To better represent the teacher model’s sequence-level distribution, we follow Section \ref{temperature_scheduled_Curriculum}, and perform two-stage SFT on Qwen3-4B-Instruct-2507: first using low-temperature sampling data, followed by high-temperature sampling data. Training configurations are kept identical across both stages. We use an initial learning rate of 5e-5 that decays to 1e-5 via a cosine scheduler. We set the cutoff length to 64K and employ greedy sequence packing to accelerate training. Given the substantial GPU memory demands of 64K-context training, we leverage ZeRO-3 optimization together with Liger kernels to reduce memory consumption. Training is conducted with a global batch size of 64 over 6 epochs, and we observe consistent performance improvements across epochs.

\subsubsection{Mixed-policy Distillation}

To mitigate the exposure bias, and inspired by the effectiveness of on-policy data \citep{chen2025retaining}, we propose a mixed-policy revision protocol.
Starting from the DAS-curated training set, we sample 50K questions. Each question is fed to the student model trained in the previous stage to generate responses. To align with the teacher’s reference length, we cap the student’s generation at 1.5 times the token count of the corresponding teacher response to the same question. Among the above student-generated solutions, we identify 15K truncated responses. For each truncated response, we discard the portion after a randomly selected position located beyond half of its total length, and then employ the teacher model to rewrite this discarded part. 
The teacher continuations that pass predefined quality filters are retained for the student’s further fine-tuning, yielding a total of mixed-policy 12.7K data samples.

%% file: sections/4_Evaluation.tex
\section{Experimental Evaluation}
\label{section-evaluaton}

\subsection{Benchmarks}

We evaluate models on five complementary reasoning benchmarks:
\begin{itemize}
    \item \textbf{AIME24}\&\textbf{AIME25} \citep{aime}: Problem sets from that year’s American Invitational Mathematics Examination (AIME) I/II, each comprising 30 challenging problems, focusing on mathematical reasoning and requiring the correct final answer.
    \item \textbf{GPQA Diamond (GPQA-D)} \citep{gpqa}: A graduate-level benchmark of 198 expert-written multiple-choice questions spanning physics, chemistry, and biology, emphasizing ''Google-proof" deep academic reasoning.
    \item \textbf{LiveCodeBench (LCB)} \citep{livecodebench}:
    A continuously updated coding benchmark that mitigates data contamination through strict temporal partitioning.
    Beyond code generation, it evaluates self-repair, executable correctness, and test-output prediction. 
    The benchmark is released in time-based snapshots; \textbf{v5} contains problems collected from October 2024 to February 2025, and \textbf{v6} covers problems collected from February 2025 to May 2025.
\end{itemize}

\subsection{Baselines}
We release our model, \name{}, and evaluate its performance on five public reasoning benchmarks: AIME24, AIME25, LCB and GPQA\mbox{-}D. We compare against two families of state-of-the-art open-source models serving as our primary baselines. All baselines were selected for their demonstrated strength in reasoning and complex problem solving, ensuring a relevant and competitive evaluation context for \name{}. Particularly, we categorize these baselines based on the accessibility of their training data. This enables a dual-perspective assessment of \name{}:
(i) against top-performing models trained on private or proprietary data, and
(ii) against leading models representing fully transparent, reproducible research.

\begin{itemize}[leftmargin=2em]
  \item \textbf{Open-Weights Only.} 
  This group comprises models that release their weights publicly but keep their training data as proprietary. These models often represent the best performance available from models where the full training process is not disclosed. Our comparison set includes: 

  \begin{itemize}
    \item The \textbf{Qwen3 series} (4B-Thinking-2507, 8B, 14B, 32B) \citep{yang2025qwen3}: The world's most popular model family featuring mainline models (8B, 14B, 32B) with switchable reasoning modes, alongside a dedicated 4B-Thinking variant optimized for complex reasoning over long contexts.   
    \item \textbf{DeepSeek-R1-0528-Qwen3-8B} \citep{guo2025deepseek}: A specialized 8B reasoning model distilled from the powerful teacher model DeepSeek-R1-0528 into a Qwen3-8B backbone.
    \item The \textbf{GLM-Z1 series} (32B-0414, 9B-0414) \citep{chatglm4}: Models built on the GLM-4 architecture,
    enhanced via extended reinforcement learning for mathematical, logical, and coding reasoning.
    \item The \textbf{Mistral 3 series} (3B, 8B) \citep{mistralai2025mistral3}: 
    Compact open-weight models that emphasize efficient reasoning, strong math and coding capabilities, and multilingual generalization, suitable for low-latency, low-memory deployments.
    
  \end{itemize}

  \item \textbf{Open-Weights \& Open-Data.} This group releases both model weights and the curated reasoning datasets, enabling full reproducibility and fostering community-wide study of reasoning acquisition. Included models are:
  \begin{itemize}
    \item \textbf{AM-thinking-v1} \citep{ji2025thinking} and \textbf{OpenThoughts3-7B} \citep{guha2025openthoughts}: These models demonstrate different open-data strategies. AM-thinking combines SFT on \textbf{2.9M} distilled examples with RL on a Qwen2.5-32B base; OpenThoughts3 achieves strong 7B-level performance via SFT alone on its publicly released \textbf{1.2M} high-quality reasoning traces.
    \item The \textbf{Pai-DistillQwen-ThoughtY series} \citep{cai2025reasoning}: 
    A set of compact models (4B, 8B) distilled from DeepSeek-R1-0528 using \textbf{365K} curated examples, accompanied by full dataset release.
    \item \textbf{POLARIS-4B-Preview} \citep{Polaris2025}: A model based on Qwen3-4B that highlights the effectiveness of scaling up RL on public data to significantly improve complex, long-context reasoning.
    \item The \textbf{Nemotron family} \citep{bercovich2025llama}: This collection includes OpenReasoning-Nemotron-7B, distilled from DeepSeek-R1-0528 on a massive \textbf{30M} example open dataset, and Nemotron-Ultra-253B, a large-scale model targeting high-end reasoning tasks.    
  \end{itemize}
\end{itemize}

\begin{figure}[tb]
\centering

\begin{subfigure}[b]{0.49\linewidth}
    \centering
    \includegraphics[width=\linewidth]{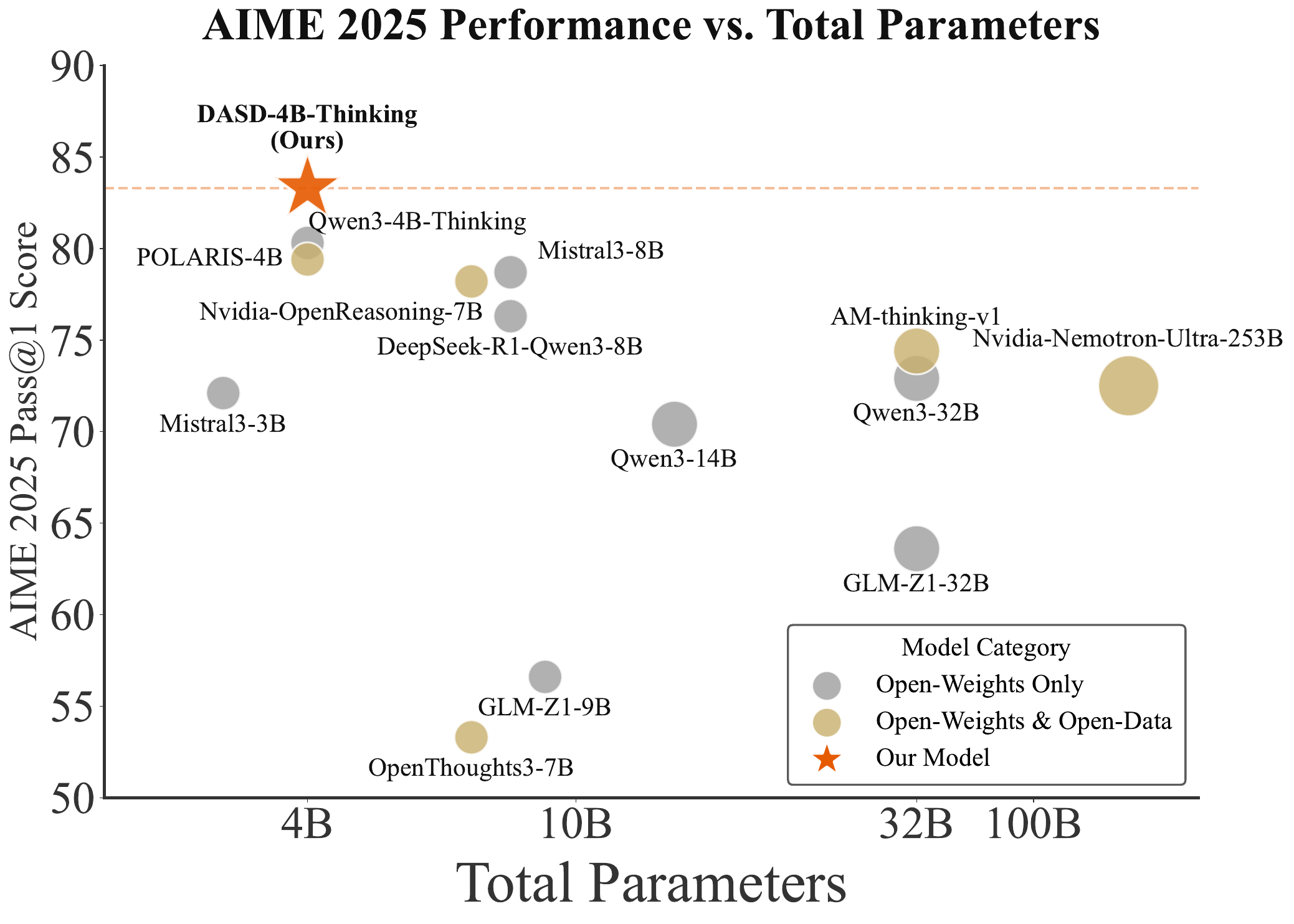}
    \label{fig:aime25}
\end{subfigure}
\begin{subfigure}[b]{0.49\linewidth}
    \centering
    \includegraphics[width=\linewidth]{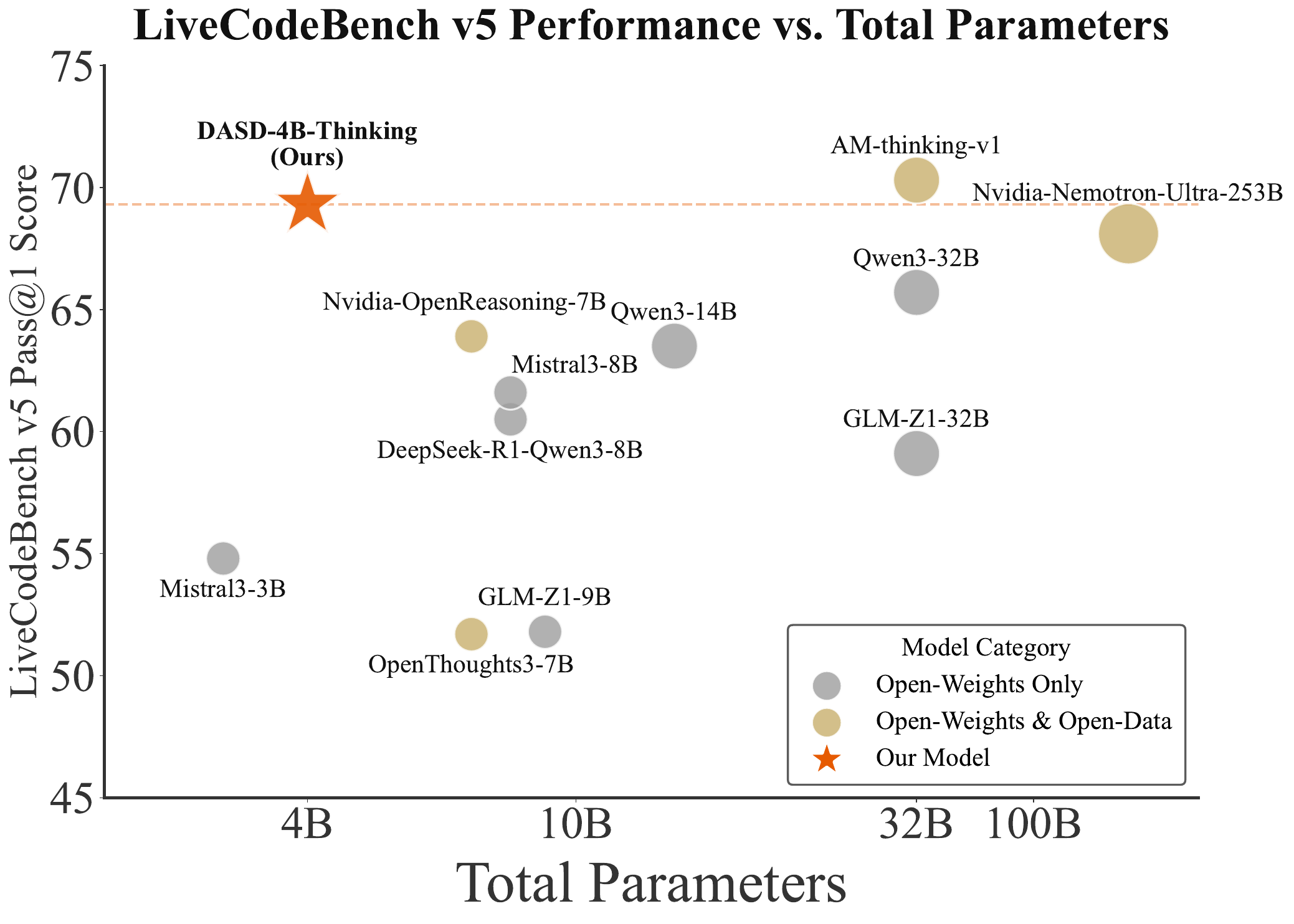}
    \label{fig:aime24}
\end{subfigure}
\vspace{-10pt}
\caption{Performance versus model size on AIME25 (left) and LCB v5 (right). Each point denotes a model, with the x-axis representing model size (in number of parameters) and the y-axis indicating benchmark score. Points positioned toward the top-left indicate superior efficiency---i.e., higher performance at smaller scale.}
\label{fig:param_vs_results}
\end{figure}

\subsection{Evaluation Setup}

All evaluations were conducted under a unified setup. We consistently set the temperature to 1.0 and top-\(p\) to \(1.0\). For every benchmark, we sampled 64 responses per question and reported the average accuracy to ensure reliable and stable evaluation results. Given the extreme difficulty of AIME24 and AIME25, we set the maximum generation length to 102{,}400 tokens; For LiveCodeBench and GPQA\mbox{-}D, the limit was set to 81{,}920 tokens.

\subsection{Main Results}

Table~\ref{tab:results} summarizes our evaluation, robustly validating the effectiveness of our proposed framework. 
The results demonstrate that, with our enhanced sequence-level distillation pipeline, complex reasoning capabilities can be efficiently transferred from a large teacher to a lightweight 4B-parameter student, yielding state-of-the-art performance for its scale. Notably, as shown in Figure~\ref{fig:param_vs_results}, \name{} not only outperforms all comparable-size models but also surpasses significantly larger counterparts (e.g., 32B) on multiple key benchmarks, highlighting both the effectiveness of our approach and the high efficiency of our training data.

\paragraph{Mathematical Reasoning (AIME24, AIME25).}
On the most challenging mathematical reasoning benchmarks, \name{} achieves remarkable scores of \textbf{83.3} on AIME25 and \textbf{88.5} on AIME24, establishing state-of-the-art performance among all listed models. These scores confirm that \name{} delivers top-tier reasoning capability, even when compared to larger scale models. This exceptional efficiency is visualized in Figure~\ref{fig:param_vs_results}, where \name{} is clearly positioned in the top-left, achieving superior performance at a fraction of the parameter cost of its competitors.

Notably, \name{}'s performance:

\begin{itemize}
    \item In the ``Open-Weights \& Open-Data'' category, our 4B model demonstrates clear superiority. It substantially outperforms the 32B AM-thinking-v1 on AIME25 (83.3 vs 74.4) and AIME24 (88.5 vs 85.3). This is particularly noteworthy given that AM-thinking-v1 was trained on 2.9M examples, whereas \name{} achieves this result using only \textbf{448K} examples---a dataset roughly 6 times smaller. It also surpasses other strong open-data models, including NVIDIA-OpenReasoning-Nemotron-7B (84.7/78.2), which used a massive 30M dataset, and even NVIDIA-Nemotron-Ultra-253B (80.8/72.5), a model over 60 times its size.
    
    \item In the ``Open-Weights Only'' category, \name{} also sets a new benchmark for compact reasoning models, decisively outperforming the strong Qwen3-4B-Thinking-2507 on AIME25 (83.3 vs. 81.3). It further exceeds several ``Open-Weights Only'' models of medium to large scale, including Qwen3-32B (81.4/72.9) and GLM-Z1-32B (80.8/63.6).
    
\end{itemize}

These results strongly demonstrate that our refined, data-efficient sequence-level distillation pipeline effectively enhances \name{}’s reasoning capabilities, enabling it to match or surpass models 8 to 60 times larger in parameter count.

\begin{table}[tb]
\centering
\caption{\textbf{Comparison across AIME24, AIME25, LiveCodeBench (v5/v6), and GPQA-D.}}
\label{tab:results}
\small
\begin{tabular}{@{}lccccc@{}}
\toprule
 & \textbf{AIME24} & \textbf{AIME25} & \textbf{LCB v5} & \textbf{LCB v6} & \textbf{GPQA-D} \\
\midrule
\multicolumn{6}{c}{\textit{Open-Weights Only}} \\
\midrule
Qwen3-4B-Thinking-2507              & -    & 81.3 & -   & 55.2   & 65.8 \\
Qwen3-14B                           & 79.3 & 70.4 & 63.5   & -   & 64.0 \\
Qwen3-32B                           & 81.4 & 72.9 & 65.7   & -   & 68.4 \\
DeepSeek\mbox{-}R1\mbox{-}0528\mbox{-}Qwen3\mbox{-}8B & 86.0 & 76.3 & 60.5   & -   & 61.1 \\
GLM\mbox{-}Z1\mbox{-}32B\mbox{-}0414 & 80.8 & 63.6 & 59.1   & -   & 66.1 \\
GLM\mbox{-}Z1\mbox{-}9B\mbox{-}0414  & 76.4 & 56.6 & 51.8   & -   & 58.5 \\
Mistral3\mbox{-}3B & - & 72.1 & 54.8   & -   & 53.4 \\
Mistral3\mbox{-}8B  & - & 78.7 & 61.6   & -   & 66.8 \\

\midrule
\multicolumn{6}{c}{\textit{Open-Weights \& Open-Data}} \\
\midrule
AM\mbox{-}thinking\mbox{-}v1                     & 85.3 & 74.4 & 70.3   & -   & - \\
POLARIS\mbox{-}4B\mbox{-}Preview                 & 81.2 & 79.4 & -   & -   & - \\
OpenThoughts3\mbox{-}7B                          & 69.0 & 53.3 & 51.7   & -   & 53.7 \\
Pai\mbox{-}DistillQwen\mbox{-}ThoughtY\mbox{-}4B & 76.7 & -    & -   & -   & 56.1 \\
Pai\mbox{-}DistillQwen\mbox{-}ThoughtY\mbox{-}8B & 76.7 & -    & -   & -   & 62.1 \\
NVIDIA\mbox{-}OpenReasoning\mbox{-}Nemotron\mbox{-}7B & 84.7 & 78.2 & 63.9   & -   & 61.4 \\
NVIDIA\mbox{-}Nemotron\mbox{-}Ultra\mbox{-}253B  & 80.8 & 72.5 & 68.1   & -   & 76.0 \\
\midrule
\rowcolor{orange!30}
\name{DASD-4B-Thinking} (Ours)                                      & \textbf{88.5} & \textbf{83.3} & \textbf{69.3} & \textbf{67.5} & \textbf{68.4} \\
\bottomrule
\end{tabular}
\end{table}

\paragraph{Coding (LiveCodeBench).}
In code generation, \name{} scores \textbf{69.3} on LCB v5 and \textbf{67.5} on LCB v6, again demonstrating exceptional efficiency.
On LCB v5, this score not only surpasses DeepSeek-R1-0528-Qwen3-8B (60.5) and Qwen3-14B (63.5), but also the NVIDIA-OpenReasoning-Nemotron-7B (63.9). On LCB v6, it outperforms the strong Qwen3-4B-Thinking-2507 (67.5 vs. 55.2).
Our 4B model even surpasses Qwen3-32B (65.7), highlighting the high efficiency of our method in transferring complex code generation capabilities.

\paragraph{Scientific QA (GPQA-D).} 

\name{} achieves a remarkable \textbf{68.4} on GPQA-D. It not only outperforms its same-size counterparts but also closely approaches the performance of substantially larger models, e.g., Qwen3-32B (68.4) and NVIDIA-Nemotron-Ultra-253B (76.0). 
While GPQA is notoriously challenging for compact models due to its heavy reliance on parametric knowledge, our 4B model successfully narrows the gap to these massive baselines. This result strongly suggests that our innovations effectively maximize the utilization of limited capacity for scientific reasoning.

\paragraph{Summary.}

Overall, \name{} delivers state-of-the-art performance across mathematical, coding, and scientific reasoning benchmarks, robustly validating the effectiveness and data efficiency of our sequence-level distillation framework. It provides new insights and a fresh perspective for developing compact, high-performing, and fully open reasoning models.

\subsection{Ablations over training stages}
\label{ablation}

Sections~\ref{temperature_scheduled_Curriculum}, \ref{divergence_aware_sampling}, and \ref{mixed_policy} have validated the individual contributions of our three core components through isolated experiments. In this section, we complement those findings with a holistic ablation of the full training pipeline, examining how performance evolves after each sequential stage. Results are summarized in Table~\ref{tab: Ablation-results}.

\begin{table}[h]
\centering
\caption{\textbf{Ablations over training stages on AIME24, AIME25, LiveCodeBench v5, LiveCodeBench v6, and GPQA-D.}}
\label{tab: Ablation-results}
\normalsize
\begin{tabular}{@{}lccccc@{}}
\toprule
 & \textbf{AIME24} & \textbf{AIME25} & \textbf{LCB v5} & \textbf{LCB v6} & \textbf{GPQA-D} \\
\midrule
Qwen3-4B-Instruct-2507 & - & 47.4 & - & 35.1 & 62.5 \\
+ Low-Temperature Training  & 84.2 & 74.0 & 56.6 & 50.6 & 67.7 \\
+ High-Temperature Training & 87.7 & 83.0 & 68.4 & 67.2 & 67.6 \\
+ Mixed-Policy Distillation   & 88.5 & 83.3 & 69.3 & 67.5 & 68.4 \\
\bottomrule
\end{tabular}
\end{table}

Starting from the Qwen3-4B-Instruct-2507 baseline, we observe consistent performance improvements across the three stages:

\noindent\textbf{Low-temperature training (with DAS)} delivers substantial initial gain, boosting AIME25 from 47.4\% to 74.0\% (+26.6\%) and LCB v6 from 35.1\% to 50.6\% (+15.5\%). This confirms that stable, low-variance gradient signals during early training are critical for establishing a solid reasoning foundation.

\noindent\textbf{High-temperature training (with DAS)} further enhances performance across key benchmarks, advancing LCB v5 by +11.8\% and LCB v6 by +16.6\%, while also providing a notable +9.0\% gain on AIME25. This demonstrates that diverse exploration under higher temperature effectively expands the policy's solution coverage once a stable baseline has been established.

\noindent\textbf{Mixed-policy distillation} consistently yields performance gains even on top of an already strong model across all benchmarks (e.g., +0.8\% on AIME24, +0.3\% on AIME25, +0.9\% on LCB v5, +0.3\% on LCB v6, +0.8\% on GPQA-D), supporting the effectiveness of mixed-policy distillation in addressing the exposure-bias issue with minimal training overhead.

\subsection{Effect of Divergence-aware Sampling on Data Distribution}

\begin{figure}[h]
  \centering
  \includegraphics[width=0.85\textwidth]{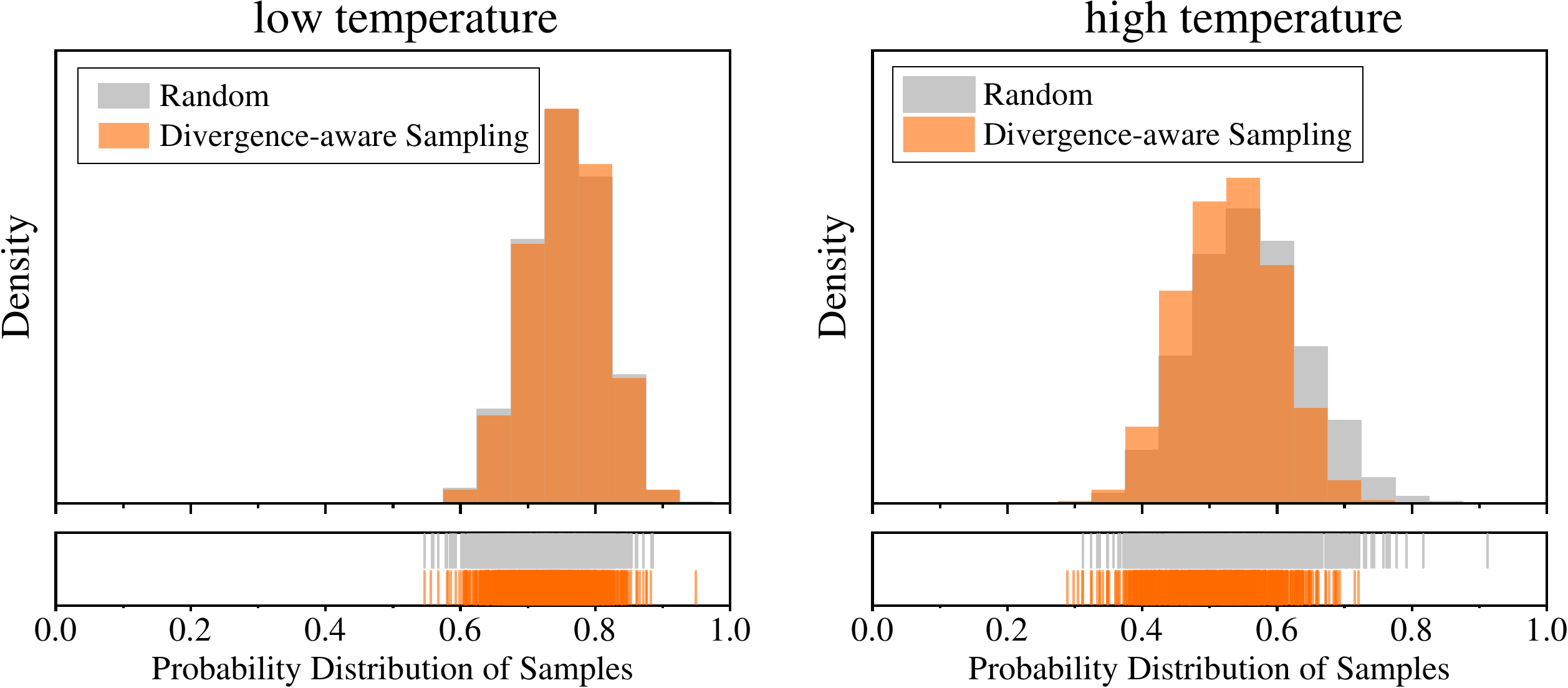}
  \caption{Comparison of response probability distributions with/without divergence-aware sampling using different temperatures.}
  \label{eval_train_data_distribution}
\end{figure}

To broaden coverage of the teacher’s output modes, we employ temperature-scheduled learning to collect both low-temperature samples (sharper, more concentrated distributions that cover a narrower probability range around high-probability regions) and high-temperature samples (flatter,  broader densities that capture rarer teacher modes). To better identify the target sequence-level distribution that supports effective student learning, we further apply divergence-aware sampling to both data subsets. As illustrated in Figure~\ref{eval_train_data_distribution}, divergence-aware sampling induces negligible perturbation to the underlying response probability distribution, confirming its orthogonality to temperature scheduling.
This decoupling underpins the strong synergy observed when combining the two strategies, as evidenced by the substantial performance gains in Section~\ref{ablation}.

\subsection{Evaluation on MoE Models}

To assess the scalability and architectural robustness of our distillation framework, we further extend it to Mixture-of-Experts (MoE) students, which increase model capacity while maintaining inference efficiency via sparse expert routing.
We select Qwen3-30B-A3B-Instruct-2507 as the student model.
For this investigation, we conduct a preliminary evaluation by applying \textbf{only the first stage} of our pipeline (i.e., Low-Temperature Training with DAS), yielding the model denoted as DASD-30B-A3B-Thinking-Preview.
Crucially, to test the cross-architecture transferability of our data, we do not re-collect or re-curate training samples; instead, we directly reuse the exact dataset curated for the Qwen3-4B student.
Table~\ref{tab:moe_results} summarizes the results. All baseline results are quoted from the corresponding official technical reports.

\begin{table}[h]
\centering
\caption{\textbf{Performance comparison on MoE models.}} 
\label{tab:moe_results}
\small
\begin{tabular}{@{}lcccc@{}}
\toprule
\textbf{Model} & \textbf{AIME25} & \textbf{LCB v6} & \textbf{GPQA-D} & \textbf{Average} \\
\midrule
gpt-oss-20b & 91.7 & 61.0* & 71.5 & 74.7 \\
Qwen3-30B-A3B-Thinking-2507 & 85.0 & 66.0 & 73.4 & 74.8 \\
NVIDIA-Nemotron-3-Nano-30B-A3B & 89.1 & 68.3 & 73.0 & 76.8 \\
\midrule
\rowcolor{orange!30}
DASD-30B-A3B-Thinking-Preview (Ours) & \textbf{86.7} & \textbf{72.8} & \textbf{72.3} & \textbf{77.3} \\
\bottomrule
\end{tabular}
\\[2pt]
{\footnotesize * indicates the number is taken from the technical report of NVIDIA-Nemotron-3.}
\end{table}

Despite being a preview version that has not been thoroughly trained, DASD-30B-A3B-Thinking-Preview already demonstrates strong competitiveness against powerful open MoE baselines. Compared with Qwen3-30B-A3B-Thinking-2507, it achieves robust gains across key benchmarks, boosting AIME25 to 86.7\% (+1.7\%) and LCB v6 to 72.8\% (+6.8\%), while maintaining competitive performance on GPQA-D. When evaluated against gpt-oss-20b, our model secures a substantial lead in coding tasks, elevating the LCB v6 score to 72.8\% (+11.8\%) and driving the overall average to 77.3\% (+2.6\%). We further compare with NVIDIA-Nemotron-3-Nano-30B-A3B \citep{nvidia_nemotron_nano_v3_2025}: according to its technical report, it is trained with a large-scale SFT corpus (18M) and additional RL, whereas our model uses only the 105K distilled dataset from the first stage of our pipeline, with no extra data or RL involved. Even under this much lighter training recipe, DASD-30B-A3B-Thinking-Preview delivers stronger coding performance on LCB v6 (+4.5; 72.8 vs.\ 68.3) and a higher average score (+0.5; 77.3 vs.\ 76.8), highlighting that careful pipeline design can achieve superior efficiency-quality trade-offs at MoE scale.

%% file: sections/5_Conclusion.tex
\section{Conclusion and Future Work}

In this technical report, we introduce DASD-4B-Thinking, a high-performance large reasoning model developed through large-scale knowledge distillation. We critically re-examine the prevailing paradigm of SFT on teacher-generated responses and identify three key limitations: (i) inadequate coverage of the teacher’s sequence-level distribution; (ii) misalignment between the teacher’s output distribution and the student’s learning capacity; and (iii) exposure bias stemming from teacher forcing during training versus autoregressive inference at test time.
To address these challenges, we propose a comprehensive data construction and training pipeline built upon three core innovations: temperature-scheduled learning, divergence-aware sampling, and mixed-policy distillation.
Leveraging this pipeline and training on only 448K samples, DASD-4B-Thinking achieves state-of-the-art performance across the majority of reasoning benchmarks---consistently outperforming models of comparable scale and, notably, surpassing several larger counterparts. Complementing this, we also release DASD-30B-A3B-Thinking-Preview, a MoE variant that attains competitive or superior results across the same benchmarks.
To promote reproducibility and community advancement, we open-source both the curated dataset and the trained models.

Looking ahead, we outline the following directions for future work. First, we will explore distribution-aware reweighting during SFT, leveraging the teacher model’s sequence-level output probabilities to more faithfully approximate its target distribution, thereby improving both distillation effectiveness and data efficiency. Second, we aim to further refine the mixed-policy distillation approach to enhance training efficiency and stability. Finally, we plan to integrate complementary agentic capabilities---such as knowledge retrieval and tool use---to progressively develop more powerful, domain-adapted reasoning models capable of handling complex, real-world tasks.

\section{Acknowledgement}

We would like to thank the following individuals (listed in alphabetical order by last name) for their valuable discussions and for their help with experiments and evaluation:

Xin Guo, Chenxi Huang, Zhen Huang, Yao Liu, Rui Miao, Ge Teng, Jingyi Shen, Xu Shen, Chaoqun Wan, Yue Xin, Xiaosong Yuan.

%% file: paper.bib
@article{hinton2015distilling,
  author       = {Geoffrey E. Hinton and
                  Oriol Vinyals and
                  Jeffrey Dean},
  title        = {Distilling the Knowledge in a Neural Network},
  journal      = {CoRR},
  volume       = {abs/1503.02531},
  year         = {2015},
}

@inproceedings{ranzato2016sequence,
  author       = {Marc'Aurelio Ranzato and
                  Sumit Chopra and
                  Michael Auli and
                  Wojciech Zaremba},
  title        = {Sequence Level Training with Recurrent Neural Networks},
  booktitle    = {International Conference on Learning Representations},
  year         = {2016},
}

@article{yang2025qwen3,
  author       = {An Yang and
                  Anfeng Li and
                  Baosong Yang and
                  Beichen Zhang and
                  Binyuan Hui and
                  Bo Zheng and
                  Bowen Yu and
                  Chang Gao and
                  Chengen Huang and
                  Chenxu Lv and
                  others},
  title        = {Qwen3 Technical Report},
  journal      = {CoRR},
  volume       = {abs/2505.09388},
  year         = {2025},
}

@inproceedings{agarwal2024on,
  author       = {Rishabh Agarwal and
                  Nino Vieillard and
                  Yongchao Zhou and
                  Piotr Stanczyk and
                  Sabela Ramos Garea and
                  Matthieu Geist and
                  Olivier Bachem},
  title        = {On-Policy Distillation of Language Models: Learning from Self-Generated Mistakes},
  booktitle    = {International Conference on Learning Representations},
  year         = {2024},
}

@inproceedings{gu2024minillm,
  author       = {Yuxian Gu and
                  Li Dong and
                  Furu Wei and
                  Minlie Huang},
  title        = {MiniLLM: Knowledge Distillation of Large Language Models},
  booktitle    = {International Conference on Learning Representations},
  year         = {2024},
}

@inproceedings{kim2016sequence,
  author       = {Yoon Kim and
                  Alexander M. Rush},
  title        = {Sequence-Level Knowledge Distillation},
  booktitle    = {Conference on Empirical Methods in Natural Language Processing},
  pages        = {1317--1327},
  year         = {2016},
}

@article{guo2025deepseek,
  author       = {Daya Guo and
                  Dejian Yang and
                  Haowei Zhang and
                  Junxiao Song and
                  Peiyi Wang and
                  Qihao Zhu and
                  Runxin Xu and
                  Ruoyu Zhang and
                  Shirong Ma and
                  Xiao Bi and
                  others},
  title        = {DeepSeek-R1 incentivizes reasoning in LLMs through reinforcement learning},
  journal      = {Nature},
  volume       = {645},
  number       = {8081},
  pages        = {633--638},
  year         = {2025},
}

@article{zhao20251,
  author       = {Han Zhao and
                  Haotian Wang and
                  Yiping Peng and
                  Sitong Zhao and
                  Xiaoyu Tian and
                  Shuaiting Chen and
                  Yunjie Ji and
                  Xiangang Li},
  title        = {1.4 Million Open-Source Distilled Reasoning Dataset to Empower Large Language Model Training},
  journal      = {CoRR},
  volume       = {abs/2503.19633},
  year         = {2025},
}

@article{guha2025openthoughts,
  author       = {Etash Kumar Guha and
                  Ryan Marten and
                  Sedrick Keh and
                  Negin Raoof and
                  Georgios Smyrnis and
                  Hritik Bansal and
                  Marianna Nezhurina and
                  Jean Mercat and
                  Trung Vu and
                  Zayne Sprague and
                  others},
  title        = {OpenThoughts: Data Recipes for Reasoning Models},
  journal      = {CoRR},
  volume       = {abs/2506.04178},
  year         = {2025},
}

@article{chen2025acereason,
  author       = {Yang Chen and
                  Zhuolin Yang and
                  Zihan Liu and
                  Chankyu Lee and
                  Peng Xu and
                  Mohammad Shoeybi and
                  Bryan Catanzaro and
                  Wei Ping},
  title        = {AceReason-Nemotron: Advancing Math and Code Reasoning through Reinforcement Learning},
  journal      = {CoRR},
  volume       = {abs/2505.16400},
  year         = {2025},
}

@article{bercovich2025llama,
  title={Llama-nemotron: Efficient reasoning models},
  author={Bercovich, Akhiad and Levy, Itay and Golan, Izik and Dabbah, Mohammad and El-Yaniv, Ran and Puny, Omri and Galil, Ido and Moshe, Zach and Ronen, Tomer and Nabwani, Najeeb and others},
  journal={arXiv preprint arXiv:2505.00949},
  year={2025}
}

@article{lei2025learning,
  author       = {Zhenyu Lei and
                  Zhen Tan and
                  Song Wang and
                  Yaochen Zhu and
                  Zihan Chen and
                  Yushun Dong and
                  Jundong Li},
  title        = {Learning from Diverse Reasoning Paths with Routing and Collaboration},
  journal      = {CoRR},
  volume       = {abs/2508.16861},
  year         = {2025},
}

@article{ye2025limo,
  author       = {Yixin Ye and
                  Zhen Huang and
                  Yang Xiao and
                  Ethan Chern and
                  Shijie Xia and
                  Pengfei Liu},
  title        = {{LIMO:} Less is More for Reasoning},
  journal      = {CoRR},
  volume       = {abs/2502.03387},
  year         = {2025},
}

@article{muennighoff2025s1,
  author       = {Niklas Muennighoff and
                  Zitong Yang and
                  Weijia Shi and
                  Xiang Lisa Li and
                  Li Fei{-}Fei and
                  Hannaneh Hajishirzi and
                  Luke Zettlemoyer and
                  Percy Liang and
                  Emmanuel J. Cand{\`{e}}s and
                  Tatsunori Hashimoto},
  title        = {s1: Simple test-time scaling},
  journal      = {CoRR},
  volume       = {abs/2501.19393},
  year         = {2025},
}

@article{wu2025beyond,
  author       = {Xiaojun Wu and
                  Xiaoguang Jiang and
                  Huiyang Li and
                  Jucai Zhai and
                  Dengfeng Liu and
                  Qiaobo Hao and
                  Huang Liu and
                  Zhiguo Yang and
                  Ji Xie and
                  Ninglun Gu and
                  Jin Yang and
                  Kailai Zhang and
                  Yelun Bao and
                  Jun Wang},
  title        = {Beyond Scaling Law: {A} Data-Efficient Distillation Framework for Reasoning},
  journal      = {CoRR},
  volume       = {abs/2508.09883},
  year         = {2025},
}

@article{jung2025prismatic,
  author       = {Jaehun Jung and
                  Seungju Han and
                  Ximing Lu and
                  Skyler Hallinan and
                  David Acuna and
                  Shrimai Prabhumoye and
                  Mostafa Patwary and
                  Mohammad Shoeybi and
                  Bryan Catanzaro and
                  Yejin Choi},
  title        = {Prismatic Synthesis: Gradient-based Data Diversification Boosts Generalization in {LLM} Reasoning},
  journal      = {CoRR},
  volume       = {abs/2505.20161},
  year         = {2025},
}

@article{li2025exploring,
  author       = {Hang Li and
                  Kaiqi Yang and
                  Yucheng Chu and
                  Hui Liu and
                  Jiliang Tang},
  title        = {Exploring Solution Divergence and Its Effect on Large Language Model Problem Solving},
  journal      = {CoRR},
  volume       = {abs/2509.22480},
  year         = {2025},
}

@article{li2025naturalthoughts,
  author       = {Yang Li and
                  Youssef Emad and
                  Karthik Padthe and
                  Jack Lanchantin and
                  Weizhe Yuan and
                  Thao Nguyen and
                  Jason Weston and
                  Shang{-}Wen Li and
                  Dong Wang and
                  Ilia Kulikov and
                  Xian Li},
  title        = {NaturalThoughts: Selecting and Distilling Reasoning Traces for General Reasoning Tasks},
  journal      = {CoRR},
  volume       = {abs/2507.01921},
  year         = {2025},
}

@misc{openr1,
    title = {Open R1: A fully open reproduction of DeepSeek-R1},
    url = {https://github.com/huggingface/open-r1},
    author = {{Hugging Face}},
    month = {January},
    year = {2025}
}

@article{kamath2025gemma,
  author       = {Aishwarya Kamath and
                  Johan Ferret and
                  Shreya Pathak and
                  Nino Vieillard and
                  Ramona Merhej and
                  Sarah Perrin and
                  Tatiana Matejovicova and
                  Alexandre Ram{\'{e}} and
                  Morgane Rivi{\`{e}}re and
                  Louis Rouillard and
                  others},
  title        = {Gemma 3 Technical Report},
  journal      = {CoRR},
  volume       = {abs/2503.19786},
  year         = {2025},
}

@article{agarwal2025gpt,
  author       = {Sandhini Agarwal and
                  Lama Ahmad and
                  Jason Ai and
                  Sam Altman and
                  Andy Applebaum and
                  Edwin Arbus and
                  Rahul K. Arora and
                  Yu Bai and
                  Bowen Baker and
                  Haiming Bao and
                  others},
  title        = {gpt-oss-120b {\&} gpt-oss-20b Model Card},
  journal      = {CoRR},
  volume       = {abs/2508.10925},
  year         = {2025},
}

@article{lu2025on,
  author = {Kevin Lu and Thinking Machines Lab},
  title = {On-Policy Distillation},
  journal = {Thinking Machines Lab: Connectionism},
  year = {2025},
  note = {https://thinkingmachines.ai/blog/on-policy-distillation},
  doi = {10.64434/tml.20251026},
}

@article{moshkov2025aimo,
  author       = {Ivan Moshkov and
                  Darragh Hanley and
                  Ivan Sorokin and
                  Shubham Toshniwal and
                  Christof Henkel and
                  Benedikt Schifferer and
                  Wei Du and
                  Igor Gitman},
  title        = {{AIMO-2} Winning Solution: Building State-of-the-Art Mathematical Reasoning Models with OpenMathReasoning dataset},
  journal      = {CoRR},
  volume       = {abs/2504.16891},
  year         = {2025},
}

@article{ahmad2025opencodereasoning,
  author       = {Wasi Uddin Ahmad and
                  Sean Narenthiran and
                  Somshubra Majumdar and
                  Aleksander Ficek and
                  Siddhartha Jain and
                  Jocelyn Huang and
                  Vahid Noroozi and
                  Boris Ginsburg},
  title        = {OpenCodeReasoning: Advancing Data Distillation for Competitive Coding},
  journal      = {CoRR},
  volume       = {abs/2504.01943},
  year         = {2025},
}

@article{cai2025reasoning,
  author       = {Wenrui Cai and
                  Chengyu Wang and
                  Junbing Yan and
                  Jun Huang and
                  Xiangzhong Fang},
  title        = {Reasoning with OmniThought: {A} Large CoT Dataset with Verbosity and Cognitive Difficulty Annotations},
  journal      = {CoRR},
  volume       = {abs/2505.10937},
  year         = {2025},
}

@misc{mattern2025synthetic,
      title={SYNTHETIC-1: Two Million Collaboratively Generated Reasoning Traces from Deepseek-R1}, 
      author={Justus Mattern and 
              Sami Jaghouar and 
              Manveer Basra and 
              Jannik Straube and 
              Matthew Di Ferrante and 
              Felix Gabriel and 
              Jack Min Ong and 
              Vincent Weisser and 
              Johannes Hagemann},
      year={2025},
      url={https://www.primeintellect.ai/blog/synthetic-1-release}, 
}

@article{li2025miromind,
  author       = {Xingxuan Li and
                  Yao Xiao and
                  Dianwen Ng and
                  Hai Ye and
                  Yue Deng and
                  Xiang Lin and
                  Bin Wang and
                  Zhanfeng Mo and
                  Chong Zhang and
                  Yueyi Zhang and
                  Zonglin Yang and
                  Ruilin Li and
                  Lei Lei and
                  Shihao Xu and
                  Han Zhao and
                  Weiling Chen and
                  Feng Ji and
                  Lidong Bing},
  title        = {MiroMind-M1: An Open-Source Advancement in Mathematical Reasoning via Context-Aware Multi-Stage Policy Optimization},
  journal      = {CoRR},
  volume       = {abs/2507.14683},
  year         = {2025},
}

@article{he2025deepmath,
  author       = {Zhiwei He and
                  Tian Liang and
                  Jiahao Xu and
                  Qiuzhi Liu and
                  Xingyu Chen and
                  Yue Wang and
                  Linfeng Song and
                  Dian Yu and
                  Zhenwen Liang and
                  Wenxuan Wang and
                  Zhuosheng Zhang and
                  Rui Wang and
                  Zhaopeng Tu and
                  Haitao Mi and
                  Dong Yu},
  title        = {DeepMath-103K: {A} Large-Scale, Challenging, Decontaminated, and Verifiable Mathematical Dataset for Advancing Reasoning},
  journal      = {CoRR},
  volume       = {abs/2504.11456},
  year         = {2025},
}

@article{wen2025light,
  author       = {Liang Wen and
                  Yunke Cai and
                  Fenrui Xiao and
                  Xin He and
                  Qi An and
                  Zhenyu Duan and
                  Yimin Du and
                  Junchen Liu and
                  Lifu Tang and
                  Xiaowei Lv and
                  Haosheng Zou and
                  Yongchao Deng and
                  Shousheng Jia and
                  Xiangzheng Zhang},
  title        = {Light-R1: Curriculum SFT, {DPO} and {RL} for Long {COT} from Scratch and Beyond},
  journal      = {CoRR},
  volume       = {abs/2503.10460},
  year         = {2025},
}

@misc{novasky2025sky,
  author       = {NovaSky Team},
  title        = {Sky-T1: Train your own O1 preview model within \$450},
  howpublished = {https://novasky-ai.github.io/posts/sky-t1},
  note         = {Accessed: 2025-01-09},
  year         = {2025}
}

@article{chen2025retaining,
  author       = {Howard Chen and
                  Noam Razin and
                  Karthik Narasimhan and
                  Danqi Chen},
  title        = {Retaining by Doing: The Role of On-Policy Data in Mitigating Forgetting},
  journal      = {CoRR},
  volume       = {abs/2510.18874},
  year         = {2025},
}

@misc{aime,
      title={{AIME} Problems and Solutions},
      author={{AIME}},
      year={2025},
      url={https://artofproblemsolving.com/wiki/index.php/AIME_Problems_and_Solutions}
}

@article{gpqa,
  author       = {David Rein and
                  Betty Li Hou and
                  Asa Cooper Stickland and
                  Jackson Petty and
                  Richard Yuanzhe Pang and
                  Julien Dirani and
                  Julian Michael and
                  Samuel R. Bowman},
  title        = {{GPQA}: A Graduate-Level {Google}-Proof {Q}{\&}{A} Benchmark},
  journal      = {CoRR},
  volume       = {abs/2311.12022},
  year         = {2023}
}

@article{livecodebench,
  author       = {Naman Jain and
                  King Han and
                  Alex Gu and
                  Wen{-}Ding Li and
                  Fanjia Yan and
                  Tianjun Zhang and
                  Sida Wang and
                  Armando Solar{-}Lezama and
                  Koushik Sen and
                  Ion Stoica},
  title        = {{LiveCodeBench}: Holistic and Contamination Free Evaluation of Large Language Models for Code},
  journal      = {CoRR},
  volume       = {abs/2403.07974},
  year         = {2024}
}

@article{chatglm4,
    title={{ChatGLM}: A Family of Large Language Models from {GLM-130B} to {GLM-4} All Tools},
    author={Aohan Zeng and 
            Bin Xu and 
            Bowen Wang and 
            Chenhui Zhang and 
            Da Yin and 
            Diego Rojas and 
            Guanyu Feng and 
            Hanlin Zhao and 
            Hanyu Lai and 
            Hao Yu and 
            others},
  journal      = {CoRR},
  volume       = {abs/2406.12793},
  year         = {2024}
}

@article{ji2025thinking,
  author       = {Yunjie Ji and
                  Xiaoyu Tian and
                  Sitong Zhao and
                  Haotian Wang and
                  Shuaiting Chen and
                  Yiping Peng and
                  Han Zhao and
                  Xiangang Li},
  title        = {AM-Thinking-v1: Advancing the Frontier of Reasoning at 32B Scale},
  journal      = {CoRR},
  volume       = {abs/2505.08311},
  year         = {2025},
}

@misc{Polaris2025,
    title = {POLARIS: A Post-Training Recipe for Scaling Reinforcement Learning on Advanced Reasoning Models},
    url = {https://hkunlp.github.io/blog/2025/Polaris},
    author = {An, Chenxin and Xie, Zhihui and Li, Xiaonan and Li, Lei and Zhang, Jun and Gong, Shansan and Zhong, Ming and Xu, Jingjing and Qiu, Xipeng and Wang, Mingxuan and Kong, Lingpeng},
    year = {2025}
}

@article{sentence_come_from,
  title={Where Did This Sentence Come From? Tracing Provenance in LLM Reasoning Distillation},
  author={Kaiyuan Liu and Shaotian Yan and Rui Miao and Bing Wang and Chen Shen and Jun Zhang and Jieping Ye},
  journal={arXiv preprint arXiv:2512.20908},
  year={2025}
}

@misc{math_dataset_numina_math_datasets,
  author = {Jia Li and Edward Beeching and Lewis Tunstall and Ben Lipkin and Roman Soletskyi and Shengyi Costa Huang and Kashif Rasul and Longhui Yu and Albert Jiang and Ziju Shen and Zihan Qin and Bin Dong and Li Zhou and Yann Fleureau and Guillaume Lample and Stanislas Polu},
  title = {NuminaMath},
  year = {2024},
  publisher = {Numina},
  journal = {Hugging Face repository},
  howpublished = {\url{[https://huggingface.co/AI-MO/NuminaMath-CoT](https://github.com/project-numina/aimo-progress-prize/blob/main/report/numina_dataset.pdf)}}
}

@article{code_dataset_taco,
  author       = {Rongao Li and
                  Jie Fu and
                  Bo{-}Wen Zhang and
                  Tao Huang and
                  Zhihong Sun and
                  Chen Lyu and
                  Guang Liu and
                  Zhi Jin and
                  Ge Li},
  title        = {{TACO:} Topics in Algorithmic COde generation dataset},
  journal      = {CoRR},
  volume       = {abs/2312.14852},
  year         = {2023},
}

@inproceedings{code_dataset_apps,
  author       = {Dan Hendrycks and
                  Steven Basart and
                  Saurav Kadavath and
                  Mantas Mazeika and
                  Akul Arora and
                  Ethan Guo and
                  Collin Burns and
                  Samir Puranik and
                  Horace He and
                  Dawn Song and
                  Jacob Steinhardt},
  title        = {Measuring Coding Challenge Competence With {APPS}},
  booktitle    = {Neural Information Processing Systems Track on Datasets and Benchmarks},
  year         = {2021},
}

@article{code_dataset_code_contest,
  author       = {Yujia Li and
                  David H. Choi and
                  Junyoung Chung and
                  Nate Kushman and
                  Julian Schrittwieser and
                  R{\'{e}}mi Leblond and
                  Tom Eccles and
                  James Keeling and
                  Felix Gimeno and
                  Agustin Dal Lago and
                  Thomas Hubert and
                  Peter Choy and
                  Cyprien de Masson d'Autume and
                  Igor Babuschkin and
                  Xinyun Chen and
                  Po{-}Sen Huang and
                  Johannes Welbl and
                  Sven Gowal and
                  Alexey Cherepanov and
                  James Molloy and
                  Daniel J. Mankowitz and
                  Esme Sutherland Robson and
                  Pushmeet Kohli and
                  Nando de Freitas and
                  Koray Kavukcuoglu and
                  Oriol Vinyals},
  title        = {Competition-Level Code Generation with AlphaCode},
  journal      = {CoRR},
  volume       = {abs/2203.07814},
  year         = {2022},
}

@misc{sci_dataset_OpenScienceReasoning2,
  author = {Nvidia},
  title = {OpenScience},
  year = {2024},
  publisher = {Nvidia},
  journal = {Hugging Face repository},
  howpublished = {\url{[https://huggingface.co/datasets/nvidia/OpenScience](https://huggingface.co/datasets/nvidia/OpenScience)}}
}

@inproceedings{holtzman2019curious,
  author       = {Ari Holtzman and
                  Jan Buys and
                  Li Du and
                  Maxwell Forbes and
                  Yejin Choi},
  title        = {The Curious Case of Neural Text Degeneration},
  booktitle    = {International Conference on Learning Representations},
  year         = {2020},
}

@inproceedings{jang2016categorical,
  author       = {Eric Jang and
                  Shixiang Gu and
                  Ben Poole},
  title        = {Categorical Reparameterization with Gumbel-Softmax},
  booktitle    = {International Conference on Learning Representations},
  year         = {2017},
}

@inproceedings{caron2021emerging,
  author       = {Mathilde Caron and
                  Hugo Touvron and
                  Ishan Misra and
                  Herv{\'{e}} J{\'{e}}gou and
                  Julien Mairal and
                  Piotr Bojanowski and
                  Armand Joulin},
  title        = {Emerging Properties in Self-Supervised Vision Transformers},
  booktitle    = {{IEEE/CVF} International Conference on Computer Vision},
  pages        = {9630--9640},
  year         = {2021},
}

@article{zhou2021ibot,
  author       = {Jinghao Zhou and
                  Chen Wei and
                  Huiyu Wang and
                  Wei Shen and
                  Cihang Xie and
                  Alan L. Yuille and
                  Tao Kong},
  title        = {iBOT: Image {BERT} Pre-Training with Online Tokenizer},
  journal      = {CoRR},
  volume       = {abs/2111.07832},
  year         = {2021},
}

@article{yan2025towards,
  author       = {Jianzhi Yan and
                  Le Liu and
                  Youcheng Pan and
                  Shiwei Chen and
                  Yang Xiang and
                  Buzhou Tang},
  title        = {Towards Efficient CoT Distillation: Self-Guided Rationale Selector for Better Performance with Fewer Rationales},
  journal      = {CoRR},
  volume       = {abs/2509.23574},
  year         = {2025},
}

@inproceedings{li2023curriculum,
  title={Curriculum temperature for knowledge distillation},
  author={Li, Zheng and Li, Xiang and Yang, Lingfeng and Zhao, Borui and Song, Renjie and Luo, Lei and Li, Jun and Yang, Jian},
  booktitle={Proceedings of the AAAI Conference on Artificial Intelligence},
  volume={37},
  number={2},
  pages={1504--1512},
  year={2023}
}

@misc{mistralai2025mistral3,
  title        = {Mistral 3},
  author       = {{Mistral AI Team}},
  year         = {2025},
  month        = {December},
  howpublished = {\url{https://mistral.ai/news/mistral-3}},
  note         = {Accessed: 2025-12-03}
}

@misc{nvidia_nemotron_nano_v3_2025,
  title  = {{Nemotron 3 Nano}: Open, Efficient Mixture-of-Experts Hybrid {Mamba}-{Transformer} Model for {Agentic} Reasoning},
  author = {{NVIDIA}},
  year   = {2025},
  url    = {https://research.nvidia.com/labs/nemotron/files/NVIDIA-Nemotron-3-Nano-Technical-Report.pdf},
  note   = {Technical report}
}
